\documentclass[11pt]{article}

\usepackage[preprint]{acl}

\usepackage{times}
\usepackage{latexsym}

\usepackage[T1]{fontenc}

\usepackage[utf8]{inputenc}
\DeclareUnicodeCharacter{2194}{\ensuremath{\leftrightarrow}}
\DeclareUnicodeCharacter{2203}{\ensuremath{\exists}}
\DeclareUnicodeCharacter{2200}{\ensuremath{\forall}}
\DeclareUnicodeCharacter{2124}{\ensuremath{\mathbb{Z}}}
\DeclareUnicodeCharacter{00AC}{\ensuremath{\neg}}
\DeclareUnicodeCharacter{2264}{\ensuremath{\le}}
\DeclareUnicodeCharacter{2227}{\ensuremath{\wedge}}
\DeclareUnicodeCharacter{2260}{\ensuremath{\ne}}
\DeclareUnicodeCharacter{22A4}{\ensuremath{\top}}

\usepackage{microtype}

\usepackage{inconsolata}

\usepackage{graphicx}

\usepackage{amsmath,amssymb}
\usepackage{booktabs,multirow,threeparttable}
\usepackage{xtab}      
\usepackage{tikz}
\usetikzlibrary{positioning,arrows.meta,shapes.geometric,fit,calc}
\usepackage{pgfplots}
\pgfplotsset{compat=1.18}
\usepackage{subcaption}
\usepackage{enumitem}
\usepackage{todonotes}

\definecolor{ofblue}{HTML}{3B6FB6}
\definecolor{oforange}{HTML}{E08A2E}
\definecolor{ofgreen}{HTML}{3E9B6D}
\definecolor{ofgrey}{HTML}{8C8C8C}
\definecolor{ofred}{HTML}{C8473E}
\definecolor{oflight}{HTML}{DDE6F2}

\newcommand{\openfrontiers}{OpenFrontiers}
\newcommand{\bench}{OpenFrontiers-Bench}
\newcommand{\redisc}{Rediscovery-110}
\newcommand{\openset}{Open-90}
\newcommand{\hq}{HQ-50}
\newcommand{\numproblems}{12,151}

\title{OpenFrontiers: A Collection of Open Research Problems\\ and a Benchmark for Scientific Rediscovery}

\author{Jiarui Yao \\
  \texttt{jiarui14@illinois.edu} \\\And
  Ruida Wang \\
  \texttt{ruidaw@illinois.edu} \\\And
  Tong Zhang \\
  \texttt{tozhang@illinois.edu}
  \AND
  \normalfont University of Illinois Urbana-Champaign}

\begin{document}
\maketitle

\begin{abstract}
    Large language models (LLMs) have shown increasing promise in research-level problem solving, yet a comprehensive collection of such problems is still missing, and evaluating them on research-level problems remains challenging. We introduce \textbf{OpenFrontiers}, a cross-disciplinary resource that combines a structured corpus of open research problems with a retrospective benchmark of recently resolved problems in mathematics. The corpus records problem statements, context, progress, and other information for 12,151 problems. The benchmark contains 200 problems, 110 resolved in 2026 and 90 still open, each with a Lean~4 statement checked by automated probes and two independent reviews; a subset \hq{} further filters based on quality score, etc. Our evaluation protocol grades formal proofs with two independent checkers in an isolated sandbox under closed-book, frozen-retrieval, and live-web conditions. Evaluating seven LLM agents on the \hq{} subset under a standardized budget, we find that closed-book rediscovery is rare, at most 7 of 40 resolved problems, whereas live-web success mostly reflects retrieval of published solutions. By linking a broad research-problem corpus to a formally graded evaluation subset with documented solutions, OpenFrontiers provides a foundation for measuring research-level rediscovery and studying progress toward solving currently open scientific problems. 
    The dataset and benchmark could be found in \url{https://huggingface.co/collections/FlippyDora/openfrontiers}.
\end{abstract}

\section{Introduction}\label{sec:intro}

Large language models (LLMs) have achieved perfect performance in extremely hard theoretical science tests, like IMO, and are now increasingly directed at problems that nobody has solved yet: systems built around them have produced formally checked proofs of olympiad problems \citep{hubert2025alphaproof} and new constructions or improved bounds for research questions \citep{romeraparedes2024funsearch,novikov2025alphaevolve,georgiev2025mathexploration}, and they appear in the solution materials of many problems examined below. Measuring the ability to resolve open problems is harder than measuring exam performance: expert-written suites such as FrontierMath \citep{glazer2024frontiermath}, Humanity's Last Exam \citep{phan2025hle} and GPQA \citep{rein2023gpqa} grade against a reference answer their authors already knew; competition sets saturate quickly and appear in training data \citep{balunovic2025matharena,sainz2023contamination,roberts2023cutoff}; and proof-level grading shows that final-answer accuracy overstates what a model has established \citep{petrov2025proofbluff}. A research problem, by contrast, is defined by the absence of a reference answer.

Measuring research-level ability therefore needs two things: a map of what is open, and tasks that can be graded at all. Open problems are scattered across problem lists, workshop reports, survey sections, prize pages, blogs and many other platforms \citep{bloom2024erdosproblems,formalconjectures2025,theoremdb2026,icmconjectures2026} with differing conventions for wording, status and licensing, and assembling them without rewording statements or overstating status is itself a curation problem. A formerly open problem can be graded only against a solution that appeared after the evaluated model was frozen, so tasks must be time-anchored \citep{white2024livebench}: the model receives the formulation that was public before resolution, and the located solution is evaluator-only.

\begin{figure*}[t]
  \centering
  \includegraphics[width=\textwidth]{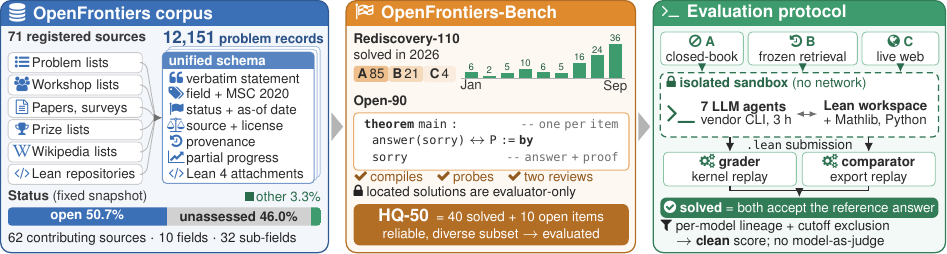}
  \caption{Overview of \openfrontiers{}: registered sources are fetched and normalized into a provenance-first corpus (Section~\ref{sec:corpus}); curation yields the 200 Lean-stated items of \bench{} (Section~\ref{sec:bench}); models are evaluated under three access conditions and graded by two Lean checkers (Section~\ref{sec:protocol}).}
  \label{fig:overview}
\end{figure*}

We introduce \openfrontiers{}, which has three coupled parts (Figure~\ref{fig:overview}). The \openfrontiers{} corpus (Section~\ref{sec:corpus}) is a large corpus of \numproblems{} problem records from 62 source collections across 10 top-level fields and 32 sub-fields, with MSC~2020 codes on 65.9\% of records and Lean~4 statements attached to 1,946. Uncertain status is labeled \emph{unassessed} (46.0\% of records) rather than promoted to \emph{open} (50.7\%).

Based on the open problems collection, we build a benchmark including problems that are still open and those resolved recently. \bench{} holds 200 research-level mathematics problems, each with a Lean~4 statement that compiles in a pinned environment and has passed automated verifications and two independent reviews. \redisc{} has 110 documented research questions that remained open into 2026 and now have a located solution, 85 with an independently checked Lean proof or exact certificate (an explicit finite object, such as a counterexample, recomputed with exact arithmetic), and \openset{} has 90 problems still open. Constructing \redisc{} yielded an observation: for 55 of the 110 the solution materials credit an AI system with the key mathematics, and most checked solutions have a public Lean proof. A model whose lineage produced a solution cannot be said to rediscover it, so post-cutoff dating alone does not control contamination, and the protocol (Section~\ref{sec:protocol}) excludes such items per model and makes closed-book the main condition.

We evaluate seven LLM agents on \hq{}, a high-quality 50-item subset of \bench{}, under closed-book, frozen-retrieval and live-web conditions, grading only Lean proofs accepted by two independent checkers. Closed-book, the agents solve 0--7 of 40 solved items, mostly three human-solved items with short constructions; with live web they solve up to 27, nearly always after fetching the published solution, and no open problem was solved without one. Frontier agents also tend to give up easily without reaching the time budget if the problem is too hard. To summarize, our contributions are:
\begin{itemize}[leftmargin=*]
  \item \textbf{Open problems corpus.} A cross-disciplinary, diverse \numproblems{} open-problem records with verbatim statements, resolve status, Lean~4 attachments, source, and other information etc.
  \item \textbf{Benchmark.} \bench{}: 200 items with checked Lean~4 statements, evidence tiers and per-item date and AI-disclosure audits, and the \hq{} subset.
  \item \textbf{Evaluation protocol.} We evaluate the models' open problem solving ability under three access conditions in an isolated sandbox, grading by two independent Lean checkers, and per-model lineage and cutoff exclusion.
  \item \textbf{Experiment results.} We conduct different evaluations on our benchmark with seven LLM agents, showing that closed-book rediscovery is rare and that live-web success reflects retrieval of published solutions.
\end{itemize}

\section{Related Work}\label{sec:related}

\paragraph{Benchmarks for mathematical and scientific reasoning.}
LLM reasoning has been tracked with competition datasets such as GSM8K \citep{cobbe2021gsm8k}, MATH \citep{hendrycks2021math}, OlympiadBench \citep{he2024olympiadbench} and Omni-MATH \citep{gao2024omnimath}, and with expert-written science questions in GPQA \citep{rein2023gpqa}, TPBench \citep{chung2025tpbench}, SciCode \citep{tian2024scicode} and LAB-Bench \citep{laurent2024labbench}. Against contamination of static benchmarks \citep{sainz2023contamination,deng2024contamination} and performance drops on post-cutoff material \citep{roberts2023cutoff}, FrontierMath \citep{glazer2024frontiermath} and Humanity's Last Exam \citep{phan2025hle} hold out expert questions, while MathArena \citep{balunovic2025matharena}, RealMath \citep{zhang2025realmath} and LiveBench \citep{white2024livebench} refresh items from new competitions and papers. Most grade a final answer or a unit-tested program on items written for the benchmark; FrontierMath: Open Problems \citep{epoch2026openproblems} and OEIS Open \citep{adamczewski2026oeisopen} pose actually open questions but accept only machine-checked answers. \openfrontiers{} also uses questions that were actually open, dates them by located solution evidence and admits expert-reviewed proofs.

\paragraph{Proof-level and formal evaluation.}
Graded as full proofs, recent olympiad problems yield few complete and rigorous arguments even from models that lead final-answer competitions \citep{petrov2025proofbluff}, motivating proof-graded sets such as IMO-ProofBench \citep{luong2025imoproofbench}. Formal benchmarks such as miniF2F \citep{zheng2021minif2f}, ProofNet \citep{azerbayev2023proofnet} and PutnamBench \citep{tsoukalas2024putnambench} replace the grader with a proof assistant, most often Lean~4 \citep{demoura2021lean4} with mathlib \citep{mathlib2020}, and dedicated provers followed: AlphaGeometry \citep{trinh2024alphageometry}, AlphaProof \citep{hubert2025alphaproof}, DeepSeek-Prover-V2 \citep{ren2025deepseekproverv2}, Kimina-Prover \citep{wang2025kimina}, Goedel-Prover \citep{lin2025goedel}, Seed-Prover \citep{chen2025seedprover} and Aristotle \citep{harmonic2025aristotle}; agentic systems such as Danus \citep{liu2026danus} go further and coordinate prover agents with an independent verifier around a memory of verified facts to attack research-level problems directly. Since only 16.0\% of corpus records carry a Lean~4 statement, \bench{} writes and reviews a pinned statement for each of its items, and our protocol grades solutions with two independent Lean checkers (Section~\ref{sec:scoring}).

\paragraph{Open-problem collections.}
Curated problem lists (the Kourovka Notebook \citep{khukhro2024kourovka}, Kirby's list \citep{kirby1997problems}, Smale's problems \citep{smale1998problems}, the Clay Millennium Prize Problems \citep{clay2000millennium}, The Open Problems Project \citep{demaine2001topp}, Ben Green's list \citep{green2024openproblems}, the AIM workshop lists \citep{aimproblemlists}), community venues (the Erd\H{o}s Problems site \citep{bloom2024erdosproblems}, the Open Problem Garden \citep{openproblemgarden}, Wikipedia's subject lists \citep{wikipedia_unsolved_math}, the COLT open-problem track \citep{colt_open_problems}) and structured databases (TheoremDB \citep{theoremdb2026}, ICM Conjectures \citep{icmconjectures2026}, Open Quantum Problems \citep{openquantumproblems}, Formal Conjectures \citep{formalconjectures2025}, the Optimization Constants \citep{davis2026optimizationconstants} and AlphaEvolve \citep{alphaevolverepository2025} repositories) are each bound to one community, one format and one notion of ``open'', and their statuses drift as problems are solved. \openfrontiers{} registers 71 such sources under one schema with verbatim statements, per-record provenance and license text, and a status policy that labels uncertain items \texttt{unassessed} rather than open (Section~\ref{sec:corpus}); Erd\H{o}s Problems and Formal Conjectures also feed \bench{} (Section~\ref{sec:bench}).

\section{The \openfrontiers{} Corpus}\label{sec:corpus}

Open problems may be gradually solved over time; therefore, we cut off the status filtering to a particular date. The corpus snapshot used here holds \numproblems{} problem records from 62 source collections in 10 top-level fields, plus 4,751 catalogue items (source indices and pointers without statements). We did not perform deduplication based on statements in order to preserve the original problems.

\subsection{Sources and acquisition}\label{sec:corpus-pipeline}

\paragraph{Source registry.}
To increase the diversity of the open-problems collection, we search as many sources as possible. The registry lists 71 sources (problem lists, workshop lists, preprints and surveys with open-question sections, prize lists, blogs and Wikipedia subject lists), of which 62 contribute problem records; the rest, such as the Kourovka Notebook \citep{khukhro2024kourovka} and Ben Green's 100 problems \citep{green2024openproblems}, are catalogue-only, with bibliographic pointers and indices collected but no statements. Seven collections have at least 100 records: TheoremDB (3,909) \citep{theoremdb2026}, the AIM Problem Lists (3,470) \citep{aimproblemlists}, ICM Conjectures (2,898) \citep{icmconjectures2026}, Erd\H{o}s Problems (634) \citep{bloom2024erdosproblems}, Formal Conjectures (459) \citep{formalconjectures2025}, Optimization Constants (115) and Open Quantum Problems (114) \citep{openquantumproblems}; eight more, including The Open Problems Project \citep{demaine2001topp}, have between 22 and 59, and the remaining 47 (including 13 Wikipedia lists \citep{wikipedia_unsolved_math}) have at most 20 records each; all 62 are listed with license notes in Table~\ref{tab:all-sources} (Appendix~\ref{app:sources}). The \texttt{verified-colt} collection (30 records: 14 open, 16 partially solved) comes from a two-reviewer fidelity audit of the 118 COLT open-problem papers of 2003--2026 \citep{colt_open_problems}.


\paragraph{Acquisition and formulation.}
To collect the data from various websites, we developed several tools as deterministic fetchers and record checksums of retrieved files. Besides, we use a unified format to store the problems. One schema covers all problem kinds; a problem item stores the verbatim statement (Markdown with LaTeX), field, MSC~2020 codes with their mapping basis, status, source and license wording, references, provenance (import and status-as-of dates), quality flags, Lean~4 attachments and partial-progress context (94.9\% of records; Appendix~\ref{app:schema}).

\paragraph{Status policy.}
Status is one of \texttt{open}, \texttt{unassessed}, \texttt{partially-solved}, \texttt{solved}, \texttt{disputed} or \texttt{withdrawn}. Historical workshop lists (AIMPL) and backlog items default to \texttt{unassessed} because their sources assert no current status; the corpus does not claim that they are still open. Every label records the source's status claim as of the stored status-as-of date, not a local check: an \texttt{open} record has not been verified to be open, and source-side ``solved'' or ``verified'' labels are kept as \emph{claims} in \texttt{source\_status\_json}, keeping the original statement.

\paragraph{Field classification and Lean 4 statement.} To facilitate the fine-grained evaluation of models on the open problems corpus, we classify the problems according to the MSC~2020 code.
Each MSC~2020 code assignment records its basis and an evidence string: \texttt{source-map} (2,665 assignments, from the source's own subject category), \texttt{community-map} (5,241, from the record's field label) or \texttt{editorial} (297, curator-assigned). Lean~4 statements \citep{demoura2021lean4,mathlib2020} are attached from TheoremDB, the Formal Conjectures repository (pinned commit \texttt{b82b08f}, Lean~4.33.1) and the Erd\H{o}s-problem files. They are display-only: no Lean is executed at export time, lexical \texttt{sorry}/\texttt{admit}/\texttt{axiom} flags are recorded instead, and formal checking happens only in the platform sandbox (Appendix~\ref{app:lean}) and the benchmark grader (Section~\ref{sec:scoring}). 

\subsection{Corpus statistics}\label{sec:corpus-stats}

\begin{table}[t]
  \centering\footnotesize
  \setlength{\tabcolsep}{3pt}
  \begin{tabular}{@{}lr@{}}
    \toprule
    Quantity & Value \\
    \midrule
    Problem / catalogue items & \numproblems{} / 4,751 \\
    Sources registered / contributing & 71 / 62 \\
    Collections with $\geq$100 records & 7 \\
    Distinct source names & 6,294 \\
    Records from problem / workshop lists & 8,349 / 3,516 \\
    Top-level fields / sub-fields & 10 / 32 \\
    Mathematics records & 11,349 (93.4\%) \\
    Status \texttt{open} & 6,162 (50.7\%) \\
    Status \texttt{unassessed} & 5,591 (46.0\%) \\
    Status \texttt{partially-solved} & 354 (2.9\%) \\
    Status \texttt{solved}/\texttt{disputed}/\texttt{withdrawn} & 41 / 2 / 1 \\
    Records with $\geq$1 MSC 2020 code & 8,011 (65.9\%) \\
    Distinct MSC codes / top-level classes & 218 / 44 \\
    Records with a Lean~4 statement & 1,946 (16.0\%) \\
    Lean attachments / flagged as placeholder & 1,961 / 1,562 \\
    References / records with $\geq$1 & 21,739 / 99.7\% \\
    Statement words: mean / median / p90 & 65.4 / 39 / 148 \\
    Records with LaTeX math & 10,570 (87.0\%) \\
    Records with full source text & 1,507 \\
    Posed year known (range 1611--2026) & 3,389 (27.9\%) \\
    \bottomrule
  \end{tabular}
  \caption{Key figures of the corpus (problem partition unless stated; snapshot 2026-10-10; source records, not deduplicated conjectures).}
  \label{tab:corpus-summary}
\end{table}

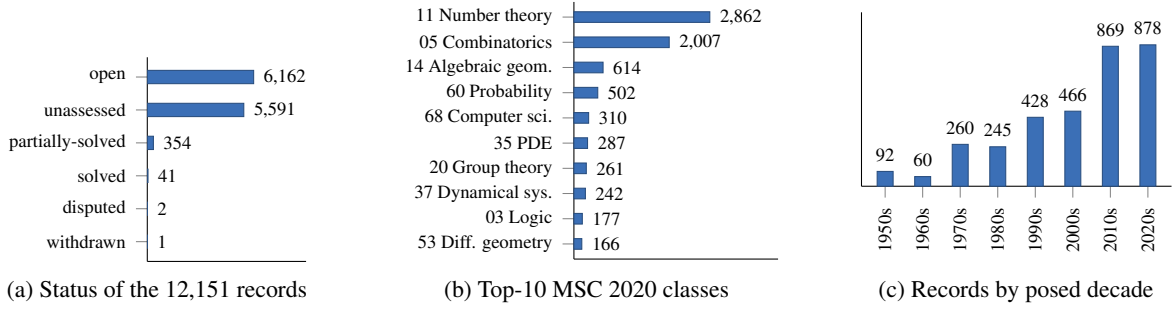
\begin{figure*}[t]
  \centering
  \begin{subfigure}[b]{0.29\linewidth}
    \centering
    \begin{tikzpicture}
      \begin{axis}[
        xbar, bar width=5pt,
        scale only axis, width=2.1cm, height=2.7cm,
        xmin=0, xmax=9200,
        xtick=\empty,
        axis x line*=bottom, axis y line*=left,
        ytick={0,1,2,3,4,5},
        yticklabels={withdrawn,disputed,solved,partially-solved,unassessed,open},
        ymin=-0.6, ymax=5.6,
        tick label style={font=\scriptsize},
        point meta=x,
        nodes near coords={\pgfmathprintnumber[fixed,precision=0,1000 sep={,},assume math mode=true]{\pgfplotspointmeta}},
        every node near coord/.append style={font=\scriptsize},
      ]
        \addplot[fill=ofblue,draw=ofblue!70!black] coordinates {
          (1,0) (2,1) (41,2) (354,3) (5591,4) (6162,5)};
      \end{axis}
    \end{tikzpicture}
    \caption{Status of the \numproblems{} records}
  \end{subfigure}\hfill
  \begin{subfigure}[b]{0.38\linewidth}
    \centering
    \begin{tikzpicture}
      \begin{axis}[
        xbar, bar width=4pt,
        scale only axis, width=2.7cm, height=3.4cm,
        xmin=0, xmax=4300,
        xtick=\empty,
        axis x line*=bottom, axis y line*=left,
        ytick={0,1,2,3,4,5,6,7,8,9},
        yticklabels={53 Diff. geometry,03 Logic,37 Dynamical sys.,20 Group theory,35 PDE,68 Computer sci.,60 Probability,14 Algebraic geom.,05 Combinatorics,11 Number theory},
        ymin=-0.6, ymax=9.6,
        tick label style={font=\scriptsize},
        point meta=x,
        nodes near coords={\pgfmathprintnumber[fixed,precision=0,1000 sep={,},assume math mode=true]{\pgfplotspointmeta}},
        every node near coord/.append style={font=\scriptsize},
      ]
        \addplot[fill=ofblue,draw=ofblue!70!black] coordinates {
          (166,0) (177,1) (242,2) (261,3) (287,4) (310,5) (502,6) (614,7) (2007,8) (2862,9)};
      \end{axis}
    \end{tikzpicture}
    \caption{Top-10 MSC 2020 classes}
  \end{subfigure}\hfill
  \begin{subfigure}[b]{0.29\linewidth}
    \centering
    \begin{tikzpicture}
      \begin{axis}[
        ybar, bar width=6pt,
        scale only axis, width=4.1cm, height=2.3cm,
        ymin=0, ymax=1080,
        ytick=\empty,
        axis x line*=bottom, axis y line*=left,
        symbolic x coords={1950s,1960s,1970s,1980s,1990s,2000s,2010s,2020s},
        xtick=data,
        enlarge x limits=0.09,
        x tick label style={font=\scriptsize,rotate=90,anchor=east},
        point meta=y,
        nodes near coords={\pgfmathprintnumber[fixed,precision=0,assume math mode=true]{\pgfplotspointmeta}},
        every node near coord/.append style={font=\scriptsize},
      ]
        \addplot[fill=ofblue,draw=ofblue!70!black] coordinates {
          (1950s,92) (1960s,60) (1970s,260) (1980s,245) (1990s,428) (2000s,466) (2010s,869) (2020s,878)};
      \end{axis}
    \end{tikzpicture}
    \caption{Records by posed decade}
  \end{subfigure}
  \caption{Corpus statistics for the \numproblems{} problem records: (a) status; (b) top-10 MSC 2020 top-level classes by records with at least one code in the class (8,011 records carry a code); (c) posed decade for the 3,389 records with a known posed year (91 records before 1950 omitted).}
  \label{fig:corpus-stats}
\end{figure*}

\paragraph{Coverage.}
Table~\ref{tab:corpus-summary} and Figure~\ref{fig:corpus-stats} summarize the status of \openfrontiers{}. Mathematics accounts for 11,349 records (93.4\%); the 802 others are mostly physics (339), computer science (285) and machine learning (70), with at most 36 records in each remaining field, so mathematics still accounts for the most formalized open problems. Within mathematics, number theory (2,867 records) and geometry and topology (2,371) lead the 32 sub-fields, and the MSC view (Figure~\ref{fig:corpus-stats}b) is led by 11~Number theory and 05~Combinatorics.

\paragraph{Status.}
Half of the records are \texttt{open} (50.7\%) and nearly as many are \texttt{unassessed} (46.0\%); the latter reflects the status policy rather than the problems, since AIMPL (3,465), TheoremDB (1,243, largely backlog candidates) and Formal Conjectures (459) account for 5,167 of them. The \texttt{open} records come mainly from TheoremDB (2,644), ICM Conjectures (2,594) and Erd\H{o}s Problems (633); 304 of the 354 \texttt{partially-solved} records come from ICM Conjectures, and only 41 are \texttt{solved}, since resolved material lives mostly in the catalogue partition (1,042 of its 4,751 items are labeled solved by their source).

\paragraph{Formal material, text and time.}
Of the 1,946 records with a Lean~4 statement, almost all from TheoremDB (1,109), Formal Conjectures (459) and Erd\H{o}s Problems (374), 1,562 are flagged \texttt{formalization-contains-placeholder} (a lexical \texttt{sorry} or \texttt{admit}), and no record carries a canonical accepted proof or a locally verified Lean proof (Appendix~\ref{app:lean}). Statements are short and most are summaries or excerpts: only 1,507 records carry the full source text. References are nearly universal (1.79 per record). Among the 3,389 records with a known posed year, the count per decade (Figure~\ref{fig:corpus-stats}c) rises from 260 in the 1970s to 869 in the 2010s, with 878 already in the 2020s.

\section{\bench{}: A Retrospective Rediscovery Benchmark}\label{sec:bench}

\bench{} is the evaluation subset of \openfrontiers{}: 200 research-level mathematics problems, each with a checked Lean~4 statement. \redisc{} holds 110 publicly posed research questions that remained open into 2026 and now have a located solution (ID prefixes R, E and N mark three curation rounds); \openset{} holds 90 problems still open at the status of check.
\hq{}, a high-quality 50-item subset selected for reference reliability and diversity (Section~\ref{sec:hq50}), is evaluated in Section~\ref{sec:protocol}. All 200 items could be found in Appendix~\ref{app:problems}.

\subsection{Construction}\label{sec:bench-design}

\paragraph{Candidates.} Solved candidates came from status changes between successive versions of the corpus’s source lists (e.g., problems newly marked solved in a later Kourovka edition, Erd\H{o}s Problems threads, Formal Conjectures \citep{formalconjectures2025}), from 2026 arXiv preprints and from public releases of AI-produced solutions; open candidates came from corpus records with an upstream Lean statement. Nine search sweeps gave 291 unique solved leads (257 verified, 230 eligible after a first pass), and 1,061 Formal Conjectures declarations were screened for the open track (152 verified, 136 kept). A second reviewer then tried to overturn each new item and rejected 5 (Appendix~\ref{app:exclusions}). Re-reviewing an earlier 90-item version removed two solved items (an answer already in the literature; a refuted solution claim) and moved four open items resolved in 2026, among them the irrationality of $\zeta(5)$, into \redisc{}.

\paragraph{Admission and evidence tiers.} An item enters \redisc{} only if it was demonstrably a public research question before its resolution, the located solution settles it with the same quantifiers and scope as posed, no earlier solution was located, and it is not a restatement of an old theorem, a withdrawn paper or an announcement without proof; same-paper and same-release items share a split group (Appendix~\ref{app:rubric}). \textbf{Tier~A} (85 items): an exact finite certificate was independently recomputed, or a complete Lean proof was compiled locally (mostly AI-assisted formalizations). \textbf{Tier~B} (21): a complete solution corroborated by the source list, a publication or an independent proof, but not re-verified by us. \textbf{Tier~C} (4): a solution claim pending verification. All located solution-evidence dates run after 2026-01-05.

\paragraph{Lean statements.} Each item states its target as one theorem \texttt{main} in a pinned environment (Lean~4.33.1, Mathlib \texttt{0df444a}, Formal Conjectures \texttt{89294ea}; \citealp{demoura2021lean4,mathlib2020}); a yes/no question is written \texttt{answer(sorry)~$\leftrightarrow$~P}, so a solution must fill in the answer \emph{and} prove \texttt{main}. 93 statements were taken verbatim from Formal Conjectures, 47 adapted (e.g., an upstream constant stronger than the posed problem) and 60 newly written; upstream problem files stay out of the build, since at least 35 of them record a solved item's answer. A checker requires every file to compile with \texttt{autoImplicit} off, without axioms or \texttt{native\_decide}, and with standard axioms only; the statement with the reference answer filled in compiles for all 110 solved items, leakage patterns (answers, upstream identifiers, live status links) are absent, and automation (\texttt{decide}, \texttt{simp}, \texttt{omega}, \texttt{aesop}) proves or refutes none of the 200 (Appendix~\ref{app:lean}).

\paragraph{Lean 4 statements review.} Two independent reviewers checked each statement, one for fidelity to the original source and one attacking its Lean semantics, with up to two repair rounds: 139 items ended ok+ok, 38 minor+ok and 23 minor+minor, with no must-fix issue. A further pass rewrote solved-item comments that hinted at the answer (47 files), keeping the comment-stripped code unchanged. These reviews were run by AI agents.

More curation criteria such as importance scores could be found in Appendix~\ref{app:rubric}.

\begin{table}[t]
\centering
\footnotesize
\setlength{\tabcolsep}{3pt}
\begin{tabular}{@{}lrrr@{}}
\toprule
\redisc{} & Jan--Jun & Jul--Sep & total \\
\midrule
Tier A / B / C & 26/8/0 & 59/13/4 & 85/21/4 \\
Erd\H{o}s Problems & 22 & 18 & 40 \\
Kourovka Notebook & 8 & 8 & 16 \\
Formal Conjectures / Wikipedia & 0 & 13 & 13 \\
Other sources (Green's list: 2) & 4 & 37 & 41 \\
AI produced the key mathematics & 14 & 41 & 55 \\
AI assistance disclosed only & 14 & 22 & 36 \\
No AI use disclosed & 6 & 13 & 19 \\
\midrule
\openset{} & & & total \\
\midrule
\multicolumn{3}{@{}l}{Formal Conjectures / Wikipedia; Erd\H{o}s; other} & 50/28/12 \\
\multicolumn{3}{@{}l}{Priority pilot / core / stretch} & 4/25/61 \\
\multicolumn{3}{@{}l}{One direction settled by a finite certificate} & 38 \\
\bottomrule
\end{tabular}
\caption{Composition of \bench{}. \redisc{} columns split by the located solution-evidence month of 2026. AI classes classify disclosures in the solution materials; ``none disclosed'' does not imply that no AI was used.}
\label{tab:bench-composition}
\end{table}

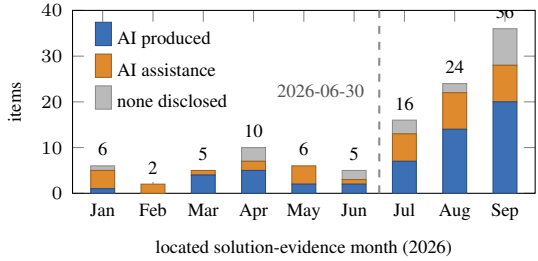
\begin{figure}[t]
\centering
\begin{tikzpicture}
\begin{axis}[
  width=\columnwidth, height=4.0cm,
  ybar stacked, bar width=9pt,
  ymin=0, ymax=40, xmin=0.4, xmax=9.6,
  xtick={1,2,3,4,5,6,7,8,9},
  xticklabels={Jan,Feb,Mar,Apr,May,Jun,Jul,Aug,Sep},
  ytick={0,10,20,30,40},
  xlabel={located solution-evidence month (2026)},
  ylabel={items},
  tick label style={font=\scriptsize},
  label style={font=\scriptsize},
  legend style={at={(0.03,0.97)}, anchor=north west, font=\scriptsize, draw=none, fill=none},
  legend cell align=left,
  enlarge x limits=false,
]
\addplot[fill=ofblue, draw=ofblue!70!black] coordinates {(1,1) (2,0) (3,4) (4,5) (5,2) (6,2) (7,7) (8,14) (9,20)};
\addplot[fill=oforange, draw=oforange!70!black] coordinates {(1,4) (2,2) (3,1) (4,2) (5,4) (6,1) (7,6) (8,8) (9,8)};
\addplot[fill=ofgrey!60, draw=ofgrey] coordinates {(1,1) (2,0) (3,0) (4,3) (5,0) (6,2) (7,3) (8,2) (9,8)};
\legend{AI produced, AI assistance, none disclosed}
\draw[dashed, thick, ofgrey] (axis cs:6.5,0) -- (axis cs:6.5,40);
\node[anchor=north east, font=\scriptsize, align=right, text=ofgrey!60!black] at (axis cs:6.4,26) {2026-06-30};
\node[above, font=\scriptsize] at (axis cs:1,6) {6};
\node[above, font=\scriptsize] at (axis cs:2,2) {2};
\node[above, font=\scriptsize] at (axis cs:3,5) {5};
\node[above, font=\scriptsize] at (axis cs:4,10) {10};
\node[above, font=\scriptsize] at (axis cs:5,6) {6};
\node[above, font=\scriptsize] at (axis cs:6,5) {5};
\node[above, font=\scriptsize] at (axis cs:7,16) {16};
\node[above, font=\scriptsize] at (axis cs:8,24) {24};
\node[above, font=\scriptsize] at (axis cs:9,36) {36};
\end{axis}
\end{tikzpicture}
\caption{Located solution-evidence month of the 110 \redisc{} items, by disclosed AI involvement (monthly totals above the bars).}
\label{fig:bench-timeline}
\end{figure}

\subsection{\redisc{} and \openset{}}\label{sec:redisc}

\paragraph{\redisc{}.} \redisc{} spans nine areas, led by group theory and number theory (19 each), graph theory (14), analysis and PDE (13) and geometry and topology (12), and includes long-standing conjectures such as Beck--Fiala, Sendov's conjecture and the cycle double cover conjecture (Table~\ref{tab:bench-composition}). 76 items have located solution evidence after 2026-06-30, so a model frozen before July can meet them unseen. Tier~A items mostly rest on a full Lean proof of the answered statement (75 of 85); the other 10 rest on an exact certificate.

\paragraph{AI involvement.} For 55 of the 110 items the solution materials credit an AI system with the key mathematics (OpenAI models 46, Anthropic 5, Google 4, Harmonic 3; several items name more than one), concentrated after June (41 of 76; Figure~\ref{fig:bench-timeline}); 36 disclose assistance with a human-attributed solution, and 19 disclose none. A model whose lineage produced a solution has in effect seen it, and most Tier~A items have a public Lean proof, so a post-cutoff date is necessary but not sufficient: the protocol excludes such items per model and makes closed-book the main condition (Section~\ref{sec:access}).

\paragraph{\openset{}.} \openset{} keeps 26 problems of the earlier open split and adds 64; number theory (25) and graph theory (20) lead, and statements come verbatim (69) or adapted (21) from Formal Conjectures. For 38 items one direction is settled by a single finite object (a counterexample or construction), so search plus reasoning could succeed. We recently checked the status to keep it relatively reliable. A kernel-accepted solution would be new mathematics, but a statement error is the likelier explanation, so any acceptance triggers review before a claim (Section~\ref{sec:scoring}).

\subsection{\hq{}}\label{sec:hq50}

\hq{} (40 solved and 10 open items) is chosen by a reproducible script in three layers. Hard filters keep statements with high formalization confidence and a clean review, solved items of unconditional Tier~A (excluding B, C and four conditional A items) with importance $\geq 4$ (human-solved post-cutoff exceptions at 3), and open items with importance $\geq 4$, clarity 5, an unrepaired statement and an answer that is not easy to guess; 55 solved and 48 open items qualify. A score then ranks them (review outcome, importance, proof type, recency), and greedy selection enforces caps: one item per paper or problem family, two per public release, at most 16 Erd\H{o}s and 6 Kourovka items and 8 per area among solved items, and at most 21 items whose key mathematics came from OpenAI models. The 40 solved items have 34 Lean-proof and 6 certificate references, 32 post-cutoff dates, and 21 OpenAI-, 2 Anthropic-, 2 Google-, and 2 other AI-produced solutions (13 human); 8 of the 10 open items have a finite-certificate direction. We rechecked all 50 statements to confirm their open status during the LLM agent evaluation.

\section{Evaluation Protocol and Results}
\label{sec:protocol}

\begin{table*}[t]
\centering
\small
\setlength{\tabcolsep}{5pt}
\begin{tabular}{@{}ll rr rr rr@{}}
\toprule
 & & \multicolumn{2}{c}{A: closed-book} & \multicolumn{2}{c}{B: frozen retrieval} & \multicolumn{2}{c}{C: live web} \\
\cmidrule(lr){3-4}\cmidrule(lr){5-6}\cmidrule(lr){7-8}
Model & lineage, cutoff & all & clean & all & clean & all & clean \\
\midrule
\multicolumn{8}{@{}l}{\emph{Native protocol (the agent ends the session)}} \\
Claude Sonnet 5.5 & Anthropic, 2026-06 & 0 & 0/30 & 1 & 1/30 & 6 & 3/30 \\
GPT-6.1-sol & OpenAI, 2026-06 & 2 & 2/15 & 2 & 2/15 & 15 & 6/15 \\
GPT-6-luna & OpenAI, 2026-06 & 0 & 0/15 & 0 & 0/15 & 4 & 3/15 \\
Kimi-K3 & --, 2026-06 & 2 & 2/33 & 2 & 2/33 & 15 & 12/33 \\
GLM-5.3 & --, 2026-08 & 2 & 2/16 & 2 & 2/16 & 15 & 8/16 \\
DeepSeek-V4.1-Flash & --, 2026-09 & 3 & 1/5 & 3 & 1/5 & 13 & 0/5 \\
Qwen3.8-27B & --, 2026-08 & 0 & 0/19 & 0 & 0/19 & 1 & 0/19$^\dagger$ \\
\midrule
\multicolumn{8}{@{}l}{\emph{Continuation protocol (resumed until the budget is spent)}} \\
Claude Sonnet 5.5 & Anthropic, 2026-06 & 4 & 3/30 & 3 & 3/30 & 24 & 18/30 \\
GPT-6.1-sol & OpenAI, 2026-06 & 7 & 4/15 & 5 & 3/15 & 27 & 12/15 \\
GPT-6-luna & OpenAI, 2026-06 & 2 & 2/15 & 0 & 0/15 & 16 & 6/15 \\
Kimi-K3 & --, 2026-06 & 3 & 3/33 & 2 & 2/33 & 15 & 13/33 \\
GLM-5.3 & --, 2026-08 & 0 & 0/16 & 2 & 2/16 & 13 & 5/16 \\
DeepSeek-V4.1-Flash & --, 2026-09 & 3 & 1/5 & 3 & 1/5 & 15 & 2/5 \\
Qwen3.8-27B & --, 2026-08 & 0 & 0/19 & 0 & 0/19 & 2 & 0/19 \\
\bottomrule
\end{tabular}

\caption{Results on the 40 solved items of \hq{} (pass@1, 3\,h per attempt). \emph{all}: items solved, i.e.\ a Lean proof with the reference answer accepted by both checkers, out of 40; \emph{clean}: solved / eligible after removing items whose key mathematics came from the model's lineage and items with solution evidence on or before the model's cutoff (release date where none is published; -- means no lineage exclusion). C mostly measures retrieval of known solutions (Section~\ref{sec:results}). Open items are reported separately.}
\label{tab:results}
\end{table*}

\subsection{Access conditions and environment}
\label{sec:access}

Each attempt runs one problem in a fresh sandbox holding only that problem's Lean workspace (with prebuilt Mathlib, so the model can compile and test its own files) and Python with numpy, sympy and mpmath. Answers, the evaluator-only package and the grading workspace are not visible; every command runs without network under an unprivileged user with private namespaces, seccomp and cgroup limits; and the problem file carries an opaque ID with status-revealing source lines removed. The three conditions differ only in retrieval, which the harness performs outside the sandbox and logs: (A)~\emph{closed-book}, no retrieval, the main condition; (B)~\emph{frozen retrieval}, search and fetch over arXiv, Wikipedia, OpenAlex and MathOverflow limited to content public before 2026-01-06, one week before the earliest solution evidence, with each hit re-checked against the historical version it displays; (C)~\emph{live web}, all URLs logged, a leakage reference rather than a rediscovery measure, since 34 of the 40 solved items have public Lean proofs; this mainly reveals the information retrieval ability of models.

\subsection{Grading}
\label{sec:scoring}

A submission is a Lean file; natural-language proofs and certificates may accompany it but are reported separately, and the main metric counts formal solutions only. Static rules admit only Mathlib imports and reject \texttt{sorry}, axioms, \texttt{native\_decide}, metaprogramming and syntax extensions; a yes/no answer must be the literal \texttt{True} or \texttt{False}. The grader compiles the reference statement with the submitted answer and the submission in separate jails, replays the submission in the kernel, compares the elaborated definitions and the statement of \texttt{main} with the reference, and allows only \texttt{propext}, \texttt{Classical.choice} and \texttt{Quot.sound}. A second, independent checker, leanprover/comparator, compares every constant the statement uses and replays the whole exported environment in the kernel without loading files the submission may have written. An item is \emph{solved} only if both checkers accept with the reference answer; a proof of the opposite answer or an accepted open item freezes the item for statement review, and a checker disagreement counts as unsolved. There is no model-as-judge. A grader self-test (accept and reject cases, including a compile-time file write that passes keyword bans) and comparator's 16 official regression attacks pass. The \emph{clean} subset removes, per model, items whose key mathematics came from its lineage and items with solution evidence on or before its cutoff; at a 2026-06-30 cutoff it holds 15 items for OpenAI models, 30 for Anthropic and 32 for unrelated models. We report counts with their denominators, never answer accuracy (a yes/no guess scores 50\%).

\subsection{Models and harness}
\label{sec:setup}

We evaluate seven models through vendor agent CLIs that run inside the sandbox behind a gateway, which pins the model and strips server-side tools in A and B: Claude Code for Claude Sonnet 5.5, Kimi-K3, GLM-5.3, DeepSeek-V4.1-Flash and Qwen3.8-27B (128k-token output limit), and Codex for GPT-6.1-sol and GPT-6-luna. Each attempt has 3 hours of wall-clock time, 5 CPU cores and 24\,GB of memory, with one attempt per item and condition. Under the \emph{native} protocol the session ends when the agent stops. Because frontier agents often stop within minutes, the \emph{continuation} protocol resumes the same session, stating the remaining budget, until a submission passes the pre-submission check or the budget is spent; an attempt ends as stalled after 20 consecutive (or 30 of 40) rounds without tool use (Appendix~\ref{app:eval}).

\subsection{Results}
\label{sec:results}

\paragraph{Closed-book rediscovery is rare.} Models solve 0--3 of the 40 items in A under the native protocol and 0--7 under continuation (Table~\ref{tab:results}). Continuation helps most the agents that stop earliest: Claude Sonnet 5.5 rises from 0 to 4, GPT-6.1-sol from 2 to 7 and GPT-6-luna from 0 to 2 (clean 3/30, 4/15 and 2/15), while the open-weight models change by at most two items. Across the 14 runs (7 models $\times$ 2 running protocols), the 53 A and B solves fall on eight items, and the three importance-3 items of \hq{}, all solved by humans through short constructions (R01, R04, R29), account for 44 of them; of the 21 importance-5 items only one (N28, settled by a finite certificate) was solved, once. Frozen retrieval did not help systematically: B exceeds A in two runs (by one and two items) and trails it in three, consistent with a snapshot that predates every solution.

\paragraph{Live web measures retrieval.} In C models solve 1--27 items, but 167 of the 181 C solves came from attempts that fetched a located solution source (preprint, repository or problem page), often with a public Lean proof to adapt; the C--A gap thus estimates how retrievable the answers are. For example, Erd\H{o}s Problem~\#152 (E03), solved by a DeepMind prover agent with a public Lean proof since 2026-04-03, was solved in none of the 28 A and B attempts but in 11 of the 14 C runs, each after fetching the problem page or its forum thread, which links to that proof. This becomes pure solution-seeking rather than scientific rediscovery. However, even in the live web setting, none of the models can solve all 40 already solved problems perfectly, showing that there is still room to improve models' information search and retrieval ability.

\paragraph{Agents stop early.} Natively, Claude Sonnet 5.5 and the GPT models end most attempts within minutes (median 0.3--4.1 of 180 minutes), whereas the open-weight models run longer, Qwen3.8-27B mostly until the limit. Under continuation most agents use nearly the full budget, but most Claude Sonnet 5.5 attempts stall, repeating that no progress is possible. By Table~\ref{tab:results}, under the continuation evaluation protocol, agents turn out to solve more problems compared to the native one. Therefore even models are equipped with the intrinsic ability to solve some open problems, it should be explicitly induced by appropriate prompting and harness design.

\section{Conclusion}\label{sec:conclusion}

We presented \openfrontiers{}, a systematic collection of \numproblems{} research-level open-problem records from 62 source collections across 10 fields, and \bench{}, 200 mathematics problems with checked Lean~4 statements, graded by two independent Lean checkers under three access conditions. Upon our evaluation, seven LLM agents rarely rediscover recent solutions under the closed-book setting, and their live-web successes mostly retrieve published solutions. Even with access to the whole Internet, models do not correctly resolve all already solved problems, revealing room for better information search and retrieval ability. Moreover, although models can solve some open problems within a fixed budget, they tend to give up early before making substantive progress, highlighting the need to develop more advanced harnesses that keep the model thinking deeper and longer rather than failing to explore the open problem thoroughly.





\bibliography{ref}

\appendix
\renewcommand{\dbltopfraction}{0.85}
\renewcommand{\dblfloatpagefraction}{0.9}
\section{Benchmark item lists}\label{app:problems}

Tables~\ref{tab:items-solved} and~\ref{tab:items-open} list all 200 items of \bench{}; a bullet after the ID marks the 50 items of \hq{}. The problem column is the English item title, truncated where long; it is a curator title, not the text shown to a model, which receives the Lean file of the solver packet (Appendix~\ref{app:eval}). Areas are a curator grouping into ten areas. For \redisc{}, \emph{T} is the evidence tier (Appendix~\ref{app:rubric}), \emph{Date} the year and month of the located solution evidence, and \emph{AI} the disclosure class of the solution materials: \emph{AI}, an AI system is credited with the key mathematics; \emph{assist}, assistance is disclosed and the solution is attributed to humans; \emph{none}, no AI use was disclosed in the reviewed materials, which does not imply that none occurred. For \openset{}, \emph{Prio.} is the implementation priority (pilot, core or stretch; not a difficulty estimate) and \emph{Cert.} marks items where one direction is settled by a single finite object. Solution URLs, split groups, Lean declaration names, scope differences, review verdicts and pending gates are kept in the package files.\\

{\scriptsize
\setlength{\tabcolsep}{2pt}
\tablefirsthead{\toprule ID & Problem & Area & T & Date & AI \\ \midrule}
\tablehead{\multicolumn{6}{@{}l}{\emph{Table~\ref{tab:items-solved} continued}}\\ \toprule ID & Problem & Area & T & Date & AI \\ \midrule}
\tabletail{\midrule \multicolumn{6}{r@{}}{\emph{continued}}\\}
\tablelasttail{\bottomrule}
\bottomcaption{\redisc{} (110 items; R, E, and N stand for three rounds of data selection with different cutoff dates). $\bullet$ = in \hq{}. T = tier; Date = located solution evidence (YY-MM); AI = disclosure class.}\label{tab:items-solved}
\begin{xtabular}{@{}l p{3.75cm} p{1.12cm} c l l@{}}
R01$\bullet$ & \raggedright Lin--Wimmer rank-minimization conjecture (rank-minimization analogue of Roth's similarity theorem) & Alg./matrix & A & 26-09 & none \tabularnewline
R02 & \raggedright Huang's weak majorization problem for singular values of Hadamard products of nonnegative matrices & Alg./matrix & A & 26-09 & none \tabularnewline
R03 & \raggedright Teschner's bondage-number conjecture & Graph th. & A & 26-09 & AI \tabularnewline
R04$\bullet$ & \raggedright Gowers U\textsuperscript{k} uncertainty conjecture & Add. comb. & A & 26-09 & none \tabularnewline
R05 & \raggedright Kourovka 21.137: if the p-th powers of a finite p-group form a subgroup, must it be powerful? & Group th. & A & 26-09 & AI \tabularnewline
R06 & \raggedright Queens each attacking at most one other queen (exact extremal value) & Comb. & B & 26-08 & none \tabularnewline
R07$\bullet$ & \raggedright Green problem 3: measure of open product-free sets & Add. comb. & A & 26-07 & assist \tabularnewline
R08 & \raggedright Planar Pompeiu / Schiffer conjecture & Geom./top. & A & 26-08 & assist \tabularnewline
R10 & \raggedright Kourovka 6.30: Is the profinite completion of a residually finite Hopfian group topologically Hopfian? & Group th. & B & 26-07 & none \tabularnewline
R12 & \raggedright Kourovka Notebook Problem 18.16 (definable endomorphisms of ordered abelian groups) & Group th. & B & 26-08 & assist \tabularnewline
R13 & \raggedright Kourovka Notebook Problem 21.87 (A. Lucchini): generating a finite group from d-generator subgroups of\ldots{} & Group th. & B & 26-08 & AI \tabularnewline
R14 & \raggedright Kourovka Notebook Problem 21.142 (embedding into groups invariably generated by a p-element and a q-element) & Group th. & B & 26-08 & AI \tabularnewline
R15 & \raggedright Kourovka Notebook Problem 12.15 (finite 2-groups in which equal normal closures force conjugacy) & Group th. & A & 26-08 & AI \tabularnewline
R16 & \raggedright Kourovka Notebook Problem 21.88 (D. MacHale): odd-order groups with k(G)/$|$G$|$ = 1/17 & Group th. & B & 26-08 & none \tabularnewline
R17 & \raggedright Open Quantum Problems 35 (seven-party case): local dimensions admitting an AME(7,d) state & TCS/logic & B & 26-08 & assist \tabularnewline
R18 & \raggedright Alfeld's conjecture on vertex interpolation by C\textsuperscript{1} cubic splines & Anal./PDE & A & 26-09 & none \tabularnewline
R19 & \raggedright Periodicity conjecture for finite-dimensional algebras (Erdmann--Skowro\'nski) & Alg./matrix & A & 26-09 & AI \tabularnewline
R20 & \raggedright Quantum Fourier Entropy--Influence (QFEI) conjecture of Bu--Garcia--Jaffe--Koh--Li & Anal./PDE & B & 26-08 & assist \tabularnewline
R21 & \raggedright Polujan--Pott conjecture on bent functions whose translation designs have a 2-transitive automorphism group & Group th. & A & 26-09 & assist \tabularnewline
R22 & \raggedright Resistance diameter of a graph versus its line graph & Graph th. & C & 26-09 & none \tabularnewline
R23$\bullet$ & \raggedright Gao's conjecture $\nu$(G) = d(G) $-$ 1 for the van Emde Boas zero-sum invariant & Add. comb. & A & 26-09 & none \tabularnewline
R24 & \raggedright BFGS with exact (first-local-minimizer) strong Wolfe line searches for arbitrary line-search constants 0 < c\textsubscript{1}\ldots{} & TCS/logic & C & 26-09 & assist \tabularnewline
R25 & \raggedright Friedgut's conjecture on influential coalitions on the continuous cube [0,1]\textsuperscript{n} & Comb. & C & 26-09 & assist \tabularnewline
R26 & \raggedright Chen--Li gap-entropy conjecture for best-arm identification (order-oblivious instance-wise bound) & TCS/logic & A & 26-09 & assist \tabularnewline
R27 & \raggedright Horn's coincidence conjecture for commuting self-maps of a compact convex set & Geom./top. & B & 26-09 & none \tabularnewline
R28 & \raggedright Audenaert--Kittaneh: can the Aujla--Bourin subadditivity inequality for concave matrix functions be\ldots{} & Alg./matrix & B & 26-09 & none \tabularnewline
R29$\bullet$ & \raggedright Bourin--Lee: optimality of the constant k/4 for an odd number of contractions & Alg./matrix & A & 26-09 & assist \tabularnewline
R30 & \raggedright Decidability of reachability for branching vector addition systems (BVAS) & TCS/logic & C & 26-07 & none \tabularnewline
E01$\bullet$ & \raggedright Erd\H{o}s Problem \#106: is f(k\textsuperscript{2}+1) = k for the maximal total side length of squares packed in a unit square? & Geom./top. & A & 26-07 & AI \tabularnewline
E02 & \raggedright Erd\H{o}s Problem \#125: does A + B have positive lower density (A, B = digits \{0,1\} in base 3, 4)? & Add. comb. & A & 26-03 & AI \tabularnewline
E03$\bullet$ & \raggedright Erd\H{o}s Problem \#152: isolated elements of the sumset of a finite Sidon set & Add. comb. & A & 26-04 & AI \tabularnewline
E04$\bullet$ & \raggedright Erd\H{o}s Problem \#193: must an infinite bounded-step walk in $\mathbb{Z}$\textsuperscript{3} contain three collinear points? & Geom./top. & A & 26-08 & assist \tabularnewline
E05 & \raggedright Erd\H{o}s Problem \#347: is there a sequence with ratio tending to 2 whose cofinite subsequences all have subset\ldots{} & Number th. & A & 26-01 & assist \tabularnewline
E06$\bullet$ & \raggedright Erd\H{o}s Problem \#358: is there an increasing integer sequence whose consecutive-block sums represent every\ldots{} & Add. comb. & A & 26-02 & assist \tabularnewline
E07 & \raggedright Erd\H{o}s Problem \#387: is there an absolute constant c > 0 such that every binomial coefficient C(n, k) with 1 $\le$\ldots{} & Number th. & B & 26-05 & assist \tabularnewline
E08 & \raggedright Erd\H{o}s Problem \#477: is there a polynomial f : $\mathbb{Z}$ $\to$ $\mathbb{Z}$ of degree at least 2 and a set A $\subseteq$ $\mathbb{Z}$ such that every\ldots{} & Number th. & A & 26-06 & AI \tabularnewline
E09 & \raggedright Erd\H{o}s Problem \#570: Ramsey number of a cycle C\textsubscript{k} versus a graph with m edges & Graph th. & B & 26-01 & none \tabularnewline
E10 & \raggedright Erd\H{o}s Problem \#633: triangles that can only be cut into a square number of congruent triangles & Number th. & B & 26-04 & AI \tabularnewline
E11 & \raggedright Erd\H{o}s Problem \#650: matching integers to distinct multiples in an interval of length 2N & Number th. & A & 26-03 & assist \tabularnewline
E12$\bullet$ & \raggedright Erd\H{o}s Problem \#865: does density 5/8 + C force three elements with all pairwise sums in the set? & Add. comb. & A & 26-06 & AI \tabularnewline
E13$\bullet$ & \raggedright Erd\H{o}s Problem \#868: bases of order 2 with many representations and minimal subbases & Number th. & A & 26-01 & assist \tabularnewline
E14$\bullet$ & \raggedright Erd\H{o}s Problem \#987 (second question): can the exponential-sum limsups satisfy A\textsubscript{k} = o(k)? & Anal./PDE & A & 26-04 & AI \tabularnewline
E15 & \raggedright Erd\H{o}s Problem \#990: does the Erd\H{o}s--Tur\'an angular discrepancy bound hold with the number of non-zero\ldots{} & Anal./PDE & A & 26-04 & AI \tabularnewline
E16 & \raggedright Erd\H{o}s Problem \#997: well-distribution of the fractional parts \{$\alpha$ p\textsubscript{n}\} along the primes & Number th. & B & 26-03 & AI \tabularnewline
E17 & \raggedright Erd\H{o}s Problem \#1044: infimum of the longest boundary of a component of a polynomial lemniscate & Geom./top. & A & 26-01 & assist \tabularnewline
E18 & \raggedright Erd\H{o}s Problem \#1138: primes in intervals of length C times the maximal prime gap & Number th. & A & 26-04 & assist \tabularnewline
E19 & \raggedright Erd\H{o}s Problem \#1148: is every large n of the form x\textsuperscript{2} + y\textsuperscript{2} $-$ z\textsuperscript{2} with max(x\textsuperscript{2}, y\textsuperscript{2}, z\textsuperscript{2}) $\le$ n? & Number th. & A & 26-03 & AI \tabularnewline
E20 & \raggedright Erd\H{o}s Problem \#1153: the Lebesgue function on a fixed subinterval & Anal./PDE & A & 26-02 & assist \tabularnewline
E21$\bullet$ & \raggedright Erd\H{o}s Problem \#1196: reciprocal-log sums of primitive sets far from the origin & Number th. & A & 26-04 & AI \tabularnewline
E22 & \raggedright Erd\H{o}s Problem \#1197: integer dilates of a set of positive measure & Number th. & A & 26-04 & assist \tabularnewline
E23 & \raggedright Kourovka Notebook Problem 14.54: number of conjugacy classes versus nilpotent subgroups & Group th. & A & 26-06 & none \tabularnewline
E24 & \raggedright Kourovka Notebook Problem 17.99: faithfulness of a germ representation of the Baumslag group & Group th. & A & 26-06 & none \tabularnewline
E25 & \raggedright Kourovka Notebook Problem 21.8: groups generated by class transpositions of equal modulus & Group th. & A & 26-04 & none \tabularnewline
E26 & \raggedright Kourovka Notebook Problem 21.10: just finite presentations of finite groups & Group th. & B & 26-05 & assist \tabularnewline
E27 & \raggedright Kourovka Notebook Problem 21.14: endomorphisms that extend along every homomorphism & Group th. & B & 26-04 & none \tabularnewline
E28 & \raggedright Kourovka Notebook Problem 21.24: power graphs of finite groups --- cograph implies chordal? & Group th. & A & 26-01 & AI \tabularnewline
E29 & \raggedright Simple groups of non-uniform exponential growth (Kourovka Notebook 21.117(a)) & Group th. & B & 26-05 & assist \tabularnewline
E30 & \raggedright Rank of Z($\langle$a,B$\rangle$) $\cap$ $\langle$a,B$\rangle'$ in extensions of elementary abelian groups (Kourovka Notebook 21.150) & Group th. & A & 26-03 & AI \tabularnewline
N01$\bullet$ & \raggedright Beck--Fiala conjecture: is the discrepancy of degree-t set systems O($\surd$t)? & Comb. & A & 26-09 & AI \tabularnewline
N02 & \raggedright Kourovka 20.76: abelian subgroups of finite p-groups and the bound p\textsuperscript{(2k)} & Group th. & B & 26-09 & assist \tabularnewline
N03 & \raggedright Tu--Deng conjecture & TCS/logic & A & 26-07 & assist \tabularnewline
N04 & \raggedright Is $\zeta$(5) irrational? & Number th. & A & 26-09 & AI \tabularnewline
N05$\bullet$ & \raggedright Erd\H{o}s's unit distance problem (Erd\H{o}s Problem \#90): is u(n) < n\textsuperscript{1+$\varepsilon$} for every $\varepsilon$ > 0 and all large n? & Geom./top. & A & 26-05 & AI \tabularnewline
N06 & \raggedright Multicolour Ramsey numbers for triangles: determine lim\textsubscript{k$\to\infty$} R(3;k)\textsuperscript{1/k} (Erd\H{o}s Problem \#183) & Graph th. & A & 26-08 & AI \tabularnewline
N07 & \raggedright Benkoski--Erd\H{o}s: is $\sigma$(n)/n bounded on weird numbers? (Erd\H{o}s Problem \#825) & Number th. & A & 26-01 & assist \tabularnewline
N08$\bullet$ & \raggedright Sendov's conjecture (Ilieff's conjecture): roots and critical points of polynomials whose zeros lie in the\ldots{} & Anal./PDE & A & 26-08 & AI \tabularnewline
N09$\bullet$ & \raggedright Hopf problem: does the 6-sphere admit a complex structure? (MathOverflow 1973) & Geom./top. & A & 26-08 & AI \tabularnewline
N10$\bullet$ & \raggedright Erd\H{o}s Problem \#1 (Erd\H{o}s--Moser): must a set of n positive integers in \{1, \ldots{}, N\} with distinct subset sums have\ldots{} & Add. comb. & A & 26-09 & AI \tabularnewline
N11 & \raggedright Erd\H{o}s--S\'os conjecture on trees (Erd\H{o}s Problem \#548) & Graph th. & A & 26-09 & AI \tabularnewline
N12$\bullet$ & \raggedright Erd\H{o}s--Simonovits rational exponents conjecture (Erd\H{o}s Problem \#571) & Graph th. & A & 26-09 & AI \tabularnewline
N13 & \raggedright Fuglede's spectral set conjecture in dimension two & Add. comb. & A & 26-07 & none \tabularnewline
N14$\bullet$ & \raggedright K\"othe's conjecture: is the sum of two nil left ideals of a ring always nil? & Alg./matrix & A & 26-09 & AI \tabularnewline
N15$\bullet$ & \raggedright Navier--Stokes Millennium Problem with smooth forcing: Clay statements (C) and (D) on $\mathbb{R}$\textsuperscript{3} and $\mathbb{R}$\textsuperscript{3}/$\mathbb{Z}$\textsuperscript{3} & Anal./PDE & A & 26-09 & AI \tabularnewline
N16$\bullet$ & \raggedright Erd\H{o}s's degeneracy conjecture for Tur\'an numbers of bipartite graphs (Erd\H{o}s Problem \#146) & Graph th. & A & 26-08 & AI \tabularnewline
N17$\bullet$ & \raggedright Crouzeix's conjecture: is $\|$p(A)$\|$ $\le$ 2$\cdot$max\textsubscript{z $\in$ W(A)} $|$p(z)$|$ for every square matrix A? & Anal./PDE & A & 26-07 & AI \tabularnewline
N18$\bullet$ & \raggedright Rockafellar's sum problem: maximal monotonicity of sums in general real Banach spaces & Anal./PDE & A & 26-09 & AI \tabularnewline
N19$\bullet$ & \raggedright Is every group sofic? (Gromov--Weiss soficity question) & Group th. & A & 26-08 & AI \tabularnewline
N20 & \raggedright Ehrhart's volume conjecture: convex bodies whose centroid is their only interior lattice point & Geom./top. & A & 26-08 & AI \tabularnewline
N21$\bullet$ & \raggedright The cycle double cover conjecture (Szekeres 1973; Seymour 1979) & Graph th. & A & 26-07 & AI \tabularnewline
N22$\bullet$ & \raggedright Rota's unimodality conjecture for the Whitney numbers of the second kind of a matroid & Comb. & A & 26-07 & assist \tabularnewline
N23$\bullet$ & \raggedright Jaeger's Petersen colouring conjecture & Graph th. & A & 26-07 & AI \tabularnewline
N24 & \raggedright Is every finite spherical set Ramsey? (Erd\H{o}s--Graham--Montgomery--Rothschild--Spencer--Straus; Graham) & Comb. & A & 26-09 & AI \tabularnewline
N25$\bullet$ & \raggedright The Pierce--Birkhoff conjecture & Alg./matrix & A & 26-09 & AI \tabularnewline
N26 & \raggedright The Jacobian conjecture (polynomial self-maps of $\mathbb{C}$\textsuperscript{$n$}, all n) & Geom./top. & A & 26-07 & assist \tabularnewline
N27$\bullet$ & \raggedright Wallace's question on countably compact cancellative topological semigroups & Geom./top. & A & 26-08 & AI \tabularnewline
N28$\bullet$ & \raggedright Stallings' equalizer conjecture: rank of the equalizer of two monomorphisms of free groups & Group th. & A & 26-04 & none \tabularnewline
N29 & \raggedright Michael's problem: continuity of characters on commutative Fr\'echet locally m-convex algebras & Geom./top. & A & 26-09 & AI \tabularnewline
N30 & \raggedright Pre-triangulated categories versus triangulated categories (is the octahedral axiom TR4 a consequence of\ldots{} & Alg./matrix & B & 26-08 & assist \tabularnewline
N31 & \raggedright The Erd\H{o}s--Simonovits compactness conjecture for finite families containing no forest & Graph th. & A & 26-08 & AI \tabularnewline
N32 & \raggedright Chv\'atal's conjecture: must some largest intersecting subfamily of an independence system be a star? & Comb. & B & 26-09 & assist \tabularnewline
N33$\bullet$ & \raggedright Albertson--Berman conjecture: induced forests in planar graphs (k/n $\ge$ 1/2) & Graph th. & A & 26-08 & assist \tabularnewline
N34$\bullet$ & \raggedright Wandering domains of entire functions with bounded forward orbit & Anal./PDE & A & 26-09 & AI \tabularnewline
N35 & \raggedright Exact exponential rate of the Cohn--Elkies linear programming bound for sphere packing & Geom./top. & A & 26-08 & AI \tabularnewline
N36$\bullet$ & \raggedright Erd\H{o}s--Herzog--Piranian (Erd\H{o}s Problem \#1041): a path of length less than 2 inside \{$|$f$|$ < 1\} joining two zeros\ldots{} & Anal./PDE & A & 26-09 & AI \tabularnewline
N37$\bullet$ & \raggedright Is $\Sigma$ a\textsubscript{n} / 2\textsuperscript{a\textsubscript{n}} irrational for every increasing integer sequence with a\textsubscript{n} / n $\to$ $\infty$? & Number th. & A & 26-07 & assist \tabularnewline
N38 & \raggedright Dirac's question for k = 4: a 4-vertex-critical graph without critical edges (Erd\H{o}s Problem \#944, case k = 4\ldots{} & Graph th. & A & 26-09 & AI \tabularnewline
N39 & \raggedright Two-colour van der Waerden numbers versus 2\textsuperscript{k}: does W(k)/2\textsuperscript{k} $\to$ $\infty$? (Erd\H{o}s 1980, conjecture (1)) & Graph th. & A & 26-08 & AI \tabularnewline
N40$\bullet$ & \raggedright Erd\H{o}s Problem \#421: is there a density-one sequence all of whose consecutive-block products are distinct? & Number th. & A & 26-07 & AI \tabularnewline
N41 & \raggedright Erd\H{o}s Problem \#126: does the number of distinct primes dividing $\prod$\textsubscript{a$\ne$b$\in$A}(a+b), minimised over n-element sets\ldots{} & Number th. & A & 26-09 & AI \tabularnewline
N42$\bullet$ & \raggedright Erd\H{o}s Problem \#123: is \{a\textsuperscript{k} b\textsuperscript{l} c\textsuperscript{m}\} d-complete for pairwise coprime a, b, c > 1? & Number th. & A & 26-07 & AI \tabularnewline
N43 & \raggedright Erd\H{o}s Problem \#119 (third question): partial sums of the maximum modulus of polynomials whose zeros are a\ldots{} & Anal./PDE & A & 26-07 & AI \tabularnewline
N44 & \raggedright Wegner's conjecture on piercing and packing of axis-parallel rectangles (Wegner 1965) & Comb. & A & 26-06 & assist \tabularnewline
N45$\bullet$ & \raggedright Mahler's problem on transcendental entire functions mapping Liouville numbers to Liouville numbers & Number th. & A & 26-09 & assist \tabularnewline
N46$\bullet$ & \raggedright Is the McEliece--Rodemich--Rumsey--Welch bound (1.4) on the rate of binary codes strict at every $\delta$? & TCS/logic & A & 26-08 & AI \tabularnewline
N47 & \raggedright Erd\H{o}s--Gimbel: does $\chi$(G) $-$ $\zeta$(G) tend to infinity for the random graph G(n, 1/2)? & Graph th. & A & 26-07 & assist \tabularnewline
N48 & \raggedright Talagrand's convexity problem: does the 10-fold sumset of a balanced compact set of Gaussian measure $\ge$ 0.99\ldots{} & Add. comb. & B & 26-05 & AI \tabularnewline
N49 & \raggedright Gr\"unbaum's hyperplane mass-equipartition problem in dimension 4: do four hyperplanes always cut a mass in $\mathbb{R}$\textsuperscript{4}\ldots{} & Comb. & A & 26-08 & assist \tabularnewline
N50 & \raggedright Shapiro's field-independence conjecture for Hurwitz sum-of-squares formulas: does the existence of an [r, s\ldots{} & Number th. & A & 26-05 & assist \tabularnewline
N51 & \raggedright Arnold's problem: algorithmic decidability of Lyapunov stability of the origin for polynomial vector fields\ldots{} & Anal./PDE & A & 26-09 & AI \tabularnewline
N52$\bullet$ & \raggedright Courtade--Kumar most informative Boolean function conjecture & TCS/logic & A & 26-09 & AI \tabularnewline
\end{xtabular}

}

{\scriptsize
\setlength{\tabcolsep}{2pt}
\tablefirsthead{\toprule ID & Problem & Area & Prio. & Cert. \\ \midrule}
\tablehead{\multicolumn{5}{@{}l}{\emph{Table~\ref{tab:items-open} continued}}\\ \toprule ID & Problem & Area & Prio. & Cert. \\ \midrule}
\tabletail{\midrule \multicolumn{5}{r@{}}{\emph{continued}}\\}
\tablelasttail{\bottomrule}
\bottomcaption{\openset{} (90 items). $\bullet$ = in \hq{}. Prio.\ = implementation priority; Cert.\ = one direction settled by a finite object.}\label{tab:items-open}
\begin{xtabular}{@{}l p{4.25cm} p{1.12cm} l c@{}}
O01 & \raggedright Erd\H{o}s Problem \#242 (Erd\H{o}s--Straus conjecture): for every n > 2, is 4/n a sum of three distinct unit fractions? & Number th. & core & yes \tabularnewline
O02$\bullet$ & \raggedright Lehmer's totient problem: is there a composite n with $\varphi$(n) $\mid$ n $-$ 1? & Number th. & core & yes \tabularnewline
O03 & \raggedright Erd\H{o}s--Moser equation: does 1\textsuperscript{k} + 2\textsuperscript{k} + $\cdots$ + (m-1)\textsuperscript{k} = m\textsuperscript{k} have any positive integer solution other than (k, m)\ldots{} & Number th. & core & yes \tabularnewline
O04 & \raggedright Kurepa's left factorial conjecture (prime version): no odd prime p divides !p = 0! + 1! + ... + (p-1)! & Number th. & core &  \tabularnewline
O05 & \raggedright Singmaster's conjecture: bounded multiplicity of entries (other than 1) in Pascal's triangle & Number th. & stretch &  \tabularnewline
O06 & \raggedright Perfect cuboid existence & Number th. & core & yes \tabularnewline
O07 & \raggedright Conway's 99-graph problem & Graph th. & pilot &  \tabularnewline
O08 & \raggedright Graceful tree conjecture (Ringel--Kotzig conjecture) & Graph th. & core &  \tabularnewline
O09 & \raggedright Sheehan's conjecture: every 4-regular Hamiltonian graph has a second Hamiltonian cycle & Graph th. & core &  \tabularnewline
O10 & \raggedright Reed's omega, delta, and chi conjecture & Graph th. & core &  \tabularnewline
O11 & \raggedright Vizing's domination conjecture & Graph th. & core &  \tabularnewline
O12 & \raggedright Union-closed sets conjecture (Frankl's conjecture) & Comb. & core &  \tabularnewline
O13 & \raggedright The 1/3--2/3 conjecture & Comb. & core &  \tabularnewline
O14 & \raggedright Barker sequences have length at most 13 (the Barker conjecture) & Comb. & pilot & yes \tabularnewline
O16 & \raggedright Hadamard conjecture: a Hadamard matrix of order 4k exists for every positive integer k & Comb. & core &  \tabularnewline
O17 & \raggedright The Gerstenhaber problem for three commuting matrices & Alg./matrix & pilot &  \tabularnewline
O18 & \raggedright Marcus--de Oliveira determinantal conjecture & Alg./matrix & core &  \tabularnewline
O20 & \raggedright Kaplansky zero divisor conjecture & Number th. & stretch &  \tabularnewline
O21 & \raggedright \v{C}ern\'y conjecture & TCS/logic & core &  \tabularnewline
O22$\bullet$ & \raggedright Scholz addition chain conjecture & TCS/logic & core & yes \tabularnewline
O24 & \raggedright Rule 30 center column not eventually periodic (Rule 30 Prize, Problem 1) & TCS/logic & stretch &  \tabularnewline
O25 & \raggedright Rational distance problem (unit square) & Geom./top. & core &  \tabularnewline
O26 & \raggedright Lonely runner conjecture & Geom./top. & stretch &  \tabularnewline
O27 & \raggedright Inscribed square problem (Toeplitz square peg problem) & Geom./top. & stretch &  \tabularnewline
O29 & \raggedright Irrationality of the Euler--Mascheroni constant & Number th. & stretch &  \tabularnewline
O30 & \raggedright Littlewood conjecture & Anal./PDE & stretch &  \tabularnewline
O31 & \raggedright Zariski Cancellation Problem for affine space in characteristic zero & Geom./top. & stretch & yes \tabularnewline
O32$\bullet$ & \raggedright Erd\H{o}s Matching Conjecture: maximum number of edges of an r-uniform hypergraph with no k pairwise disjoint\ldots{} & Comb. & stretch & yes \tabularnewline
O33$\bullet$ & \raggedright Erd\H{o}s--Szekeres "Happy Ending" conjecture: f(n) = 2\textsuperscript{(n-2)} + 1 & Geom./top. & stretch & yes \tabularnewline
O34 & \raggedright Gallai's path decomposition conjecture (Erd\H{o}s Problem \#583) & Graph th. & stretch & yes \tabularnewline
O35 & \raggedright Erd\H{o}s--Selfridge odd covering problem: is there a covering system with distinct odd moduli > 1? & Number th. & stretch & yes \tabularnewline
O36$\bullet$ & \raggedright Prime power conjecture for finite projective planes (Erd\H{o}s Problem \#723) & Comb. & stretch & yes \tabularnewline
O37 & \raggedright Maximum number of mutually unbiased bases in C\textsuperscript{6} & Comb. & stretch &  \tabularnewline
O38$\bullet$ & \raggedright Borsuk's partition problem in dimensions 4 $\le$ n $\le$ 62 (prove or disprove) & Geom./top. & pilot & yes \tabularnewline
O39 & \raggedright Ryser's conjecture: every Latin square of odd order has a transversal & Comb. & stretch & yes \tabularnewline
O40 & \raggedright Kummer--Vandiver conjecture: p does not divide the class number of $\mathbb{Q}$($\zeta$\textsubscript{p})\textsuperscript{+} & Number th. & stretch & yes \tabularnewline
O41$\bullet$ & \raggedright Sidorenko's conjecture: t(H, G) $\ge$ t(K\textsubscript{2}, G)\textsuperscript{e(H)} for every bipartite graph H & Graph th. & stretch & yes \tabularnewline
O42 & \raggedright Mazur rotations problem (Banach--Mazur rotation problem): is every separable infinite-dimensional Banach space\ldots{} & Anal./PDE & stretch &  \tabularnewline
O43 & \raggedright Erd\H{o}s Problem \#159: R(C\textsubscript{4}, K\textsubscript{n}) $\ll$ n\textsuperscript{2$-$c} for some c > 0 & Graph th. & stretch &  \tabularnewline
O44 & \raggedright Erd\H{o}s Problem \#20 (Erd\H{o}s--Rado sunflower conjecture): is f(n,k) < c\textsubscript{k}\textsuperscript{n}? & Comb. & stretch &  \tabularnewline
O45 & \raggedright Erd\H{o}s Problem \#39: an infinite Sidon set with counting function $\gg$\textsubscript{$\varepsilon$} N\textsuperscript{1/2$-\varepsilon$} & Add. comb. & stretch &  \tabularnewline
O46 & \raggedright Chowla's cosine problem (Erd\H{o}s Problem \#510): does every set A of N positive integers admit $\theta$ with $\Sigma$\textsubscript{n$\in$A}\ldots{} & Add. comb. & stretch &  \tabularnewline
O47 & \raggedright Erd\H{o}s Problem \#572: the even-cycle Tur\'an lower bound ex(n; C\textsubscript{2k}) $\gg$ n\textsuperscript{1+1/k} for every k $\ge$ 3 & Graph th. & stretch &  \tabularnewline
O48 & \raggedright The Erd\H{o}s--Hajnal conjecture: graphs with no induced copy of a fixed graph H contain a clique or an\ldots{} & Graph th. & stretch &  \tabularnewline
O49 & \raggedright Erd\H{o}s Problem \#564 (Erd\H{o}s--Hajnal--Rado): is the diagonal 3-uniform hypergraph Ramsey number R\textsubscript{3}(n) doubly\ldots{} & Comb. & stretch &  \tabularnewline
O50 & \raggedright Erd\H{o}s Problem \#714: is the K\H{o}v\'ari--S\'os--Tur\'an bound sharp, i.e. is ex(n; K\textsubscript{r,r}) $\gg$ n\textsuperscript{2-1/r} for every r $\ge$ 2? & Graph th. & stretch &  \tabularnewline
O51 & \raggedright Green's Problem 94 (Erd\H{o}s similarity problem for \{1, 1/2, 1/4, ...\}): does every set A $\subset$ $\mathbb{R}$ of positive\ldots{} & Anal./PDE & stretch &  \tabularnewline
O52 & \raggedright Power-saving bound for 4-AP-free sets in F\textsubscript{5}\textsuperscript{n} (Green's Problem 9(iii)) & Add. comb. & stretch &  \tabularnewline
O53$\bullet$ & \raggedright The Andrews--Curtis conjecture for balanced presentations of the trivial group & Group th. & stretch &  \tabularnewline
O54 & \raggedright Babai's conjecture for alternating groups: diam(A\textsubscript{n}) is polynomially bounded in n & Group th. & stretch &  \tabularnewline
O55 & \raggedright Falconer's distance set conjecture in the plane & Geom./top. & stretch &  \tabularnewline
O56$\bullet$ & \raggedright Lehmer's problem: Mahler measures of integer polynomials are bounded away from 1 & Number th. & stretch &  \tabularnewline
O57 & \raggedright Auslander--Reiten conjecture for Artin algebras & Alg./matrix & stretch &  \tabularnewline
O58 & \raggedright Erd\H{o}s Problem \#1029: does the diagonal Ramsey number satisfy R(k) / (k $\cdot$ 2\textsuperscript{k/2}) $\to$ $\infty$? & Graph th. & stretch &  \tabularnewline
O59 & \raggedright The Alon--Tarsi 7/5 short cycle cover conjecture & Graph th. & stretch & yes \tabularnewline
O60 & \raggedright Friedland's conjecture on the generic rank of complex 3-tensors in the critical range & Alg./matrix & stretch &  \tabularnewline
O61 & \raggedright Erd\H{o}s--Rousseau sparse-halves problem (Erd\H{o}s Problem \#128): if every induced subgraph on at least $\lfloor$n/2$\rfloor$\ldots{} & Graph th. & stretch & yes \tabularnewline
O62 & \raggedright Herzog--Sch\"onheim conjecture: in a nontrivial coset partition of a group, two subgroups have the same index\ldots{} & Group th. & stretch & yes \tabularnewline
O63 & \raggedright Erd\H{o}s Problem \#23: can every triangle-free graph on 5n vertices be made bipartite by deleting at most n\textsuperscript{2}\ldots{} & Graph th. & stretch & yes \tabularnewline
O64 & \raggedright Erd\H{o}s Problem \#364 (Erd\H{o}s; Mollin--Walsh): are there three consecutive powerful numbers? & Number th. & stretch & yes \tabularnewline
O65 & \raggedright Erd\H{o}s Problem \#324: is there an integer polynomial f whose pairwise sums f(a) + f(b), a < b nonnegative\ldots{} & Number th. & stretch &  \tabularnewline
O66 & \raggedright Do odd weird numbers exist? (Erd\H{o}s Problem \#470, first question) & Number th. & stretch & yes \tabularnewline
O67 & \raggedright Erd\H{o}s--Lov\'asz Tihany conjecture: every graph with $\omega$(G) < $\chi$(G) = a + b $-$ 1 (a, b $\ge$ 2) is (a, b)-splittable & Graph th. & stretch & yes \tabularnewline
O68 & \raggedright Erd\H{o}s--Gy\'arf\'as conjecture: every finite graph with minimum degree $\ge$ 3 contains a cycle of length 2\textsuperscript{k} (k $\ge$ 2) & Graph th. & core & yes \tabularnewline
O69 & \raggedright Green's Problem 15: is there a Lipschitz function f : $\mathbb{N}$ $\to$ $\mathbb{Z}$ whose graph contains no 3-term arithmetic\ldots{} & Graph th. & core & yes \tabularnewline
O70 & \raggedright The unfair 0-1 polynomial conjecture: if monic real polynomials P, Q with nonnegative coefficients have a\ldots{} & Probability & core & yes \tabularnewline
O71 & \raggedright Polynomial bijection from $\mathbb{Q}$ $\times$ $\mathbb{Q}$ to $\mathbb{Q}$ (MathOverflow 21003) & Number th. & stretch & yes \tabularnewline
O72$\bullet$ & \raggedright Kourovka 8.8(b) (Anosov): a non-cyclic finitely presented group in which every element is conjugate to a\ldots{} & Group th. & stretch & yes \tabularnewline
O73 & \raggedright Graham's pebbling conjecture: $\pi$(G $\square$ H) $\le$ $\pi$(G)$\cdot\pi$(H) for finite connected graphs & Graph th. & stretch & yes \tabularnewline
O74 & \raggedright Existence of a 3$\times$3 magic square of nine distinct squares (LaBar--Gardner problem) & Number th. & stretch & yes \tabularnewline
O75 & \raggedright Gottschalk's surjunctivity conjecture: is every group surjunctive? & Group th. & stretch &  \tabularnewline
O76 & \raggedright No five collinear points implies o(n\textsuperscript{2}) four-point lines (Erd\H{o}s Problem \#101) & Geom./top. & stretch &  \tabularnewline
O77 & \raggedright Green's Problem 53: if F\textsubscript{2}\textsuperscript{n} is partitioned into K sets A\textsubscript{1}, ..., A\textsubscript{K}, does some 2A\textsubscript{i} contain a coset of\ldots{} & Add. comb. & core &  \tabularnewline
O78 & \raggedright Green's Problem 21 (Rado's boundedness conjecture): is the least number of colours c(a\textsubscript{1},\ldots{},a\textsubscript{k}) needed to\ldots{} & Add. comb. & stretch &  \tabularnewline
O79 & \raggedright Meir--Moser rectangle packing: can the rectangles 1/n $\times$ 1/(n+1) (n $\ge$ 1) be packed into the unit square? & Geom./top. & stretch &  \tabularnewline
O80 & \raggedright Erd\H{o}s Problem \#41: does every infinite B\textsubscript{3} set A $\subseteq$ $\mathbb{N}$ satisfy liminf $|$A $\cap$ \{1, \ldots{}, N\}$|$ / N\textsuperscript{1/3} = 0? & Number th. & stretch &  \tabularnewline
O81 & \raggedright Quadratic sensitivity conjecture: is bs(f) = O(s(f)\textsuperscript{2})? & Comb. & stretch & yes \tabularnewline
O82 & \raggedright Jacobson's conjecture: in a left and right Noetherian ring, $\bigcap$\textsubscript{$n$} J(R)\textsuperscript{$n$} = 0 & Alg./matrix & stretch &  \tabularnewline
O83 & \raggedright Four exponentials conjecture & Number th. & stretch &  \tabularnewline
O84 & \raggedright Rudin's conjecture: does every N-term arithmetic progression contain only O($\surd$N) perfect squares? & Number th. & stretch &  \tabularnewline
O85 & \raggedright Atiyah--Sutcliffe Conjecture 1: the Atiyah configuration polynomials of n distinct points in R\textsuperscript{3} are linearly\ldots{} & Geom./top. & stretch & yes \tabularnewline
O86 & \raggedright Etingof--Ostrik finite generation conjecture (Hopf algebra case): for a finite-dimensional Hopf algebra A\ldots{} & Alg./matrix & stretch &  \tabularnewline
O87 & \raggedright Brennan's conjecture: the universal integral means spectrum satisfies B($-$2) = 1 & Geom./top. & stretch &  \tabularnewline
O88 & \raggedright Loh's 2-increasing triples problem: the Gowers--Long conjecture F(n) $\le$ n\textsuperscript{3/2} & Graph th. & core & yes \tabularnewline
O89 & \raggedright Amdeberhan--Medina--Moll conjecture: tan(arctan 1 + arctan 2 + ... + arctan n) is never an integer for n $\ge$ 5 & Number th. & core & yes \tabularnewline
O90 & \raggedright Erd\H{o}s Problem \#66: is there A $\subseteq$ $\mathbb{N}$ such that 1\textsubscript{A} $*$ 1\textsubscript{A}(n) / log n tends to a nonzero limit? & Number th. & stretch &  \tabularnewline
O91 & \raggedright Sun's conjecture: the truncations of $\sum$ C(2k,k)\textsuperscript{2} x\textsuperscript{k} and of $\sum$ C(2k,k)\textsuperscript{2} x\textsuperscript{k}/(k+1) are irreducible over $\mathbb{Q}$ & Number th. & core & yes \tabularnewline
O92 & \raggedright Sun's conjecture: every prime p has a primitive root g < p of the form k\textsuperscript{2} + 1 & Number th. & core & yes \tabularnewline
O93 & \raggedright Sun's (p4, p5, p6) conjecture: every natural number is a square plus a pentagonal number plus a hexagonal\ldots{} & Number th. & core & yes \tabularnewline
O94 & \raggedright The Goormaghtigh conjecture: 31 and 8191 are the only numbers that are repunits with at least three digits in\ldots{} & Number th. & stretch & yes \tabularnewline
\end{xtabular}

}

\section{Source collections}\label{app:sources}

Table~\ref{tab:all-sources} lists the 62 source collections that contribute problem records in the snapshot of our \openfrontiers{}, sorted by record count, with a shortened form of the reuse note recorded in the source registry (\texttt{metadata/sources.json}); the full wording travels with every item in \texttt{source.license}. The other registered sources (71 are registered in total) are catalogue-only: for the Kourovka Notebook \citep{khukhro2024kourovka} (1,935 catalogue entries), Ben Green's list of 100 problems \citep{green2024openproblems}, the K3 low-dimensional topology list and similar indices, only bibliographic pointers were collected, not statements, so they appear in the \texttt{catalogues} configuration and not in this table. Record counts are source records, not deduplicated conjectures: several collections describe the same problem, and one record may contain several sub-questions. Seven collections have at least 100 records; 47 collections have at most 20 records each. For the AlphaEvolve repository, 8 of its 67 pages (rediscovery tests of classical results, a competition problem, a programming game and a logic puzzle) are kept as catalogue rows only; the Optimization Constants and FrontierMath: Open Problems collections are imported in full. \\

{\footnotesize
\setlength{\tabcolsep}{3pt}
\tablefirsthead{\toprule Source collection & Records & License / reuse note \\ \midrule}
\tablehead{\multicolumn{3}{@{}l}{\emph{Table~\ref{tab:all-sources} continued}}\\ \toprule Source collection & Records & License / reuse note \\ \midrule}
\tabletail{\midrule \multicolumn{3}{r@{}}{\emph{continued}}\\}
\tablelasttail{\midrule Total (62 collections) & 12,151 & \\ \bottomrule}
\bottomcaption{All 62 source collections contributing problem records, sorted by record count. Records are source records, not deduplicated conjectures. The reuse note is a shortened form of the registry wording carried on each row in \texttt{source.license}; it is not a license grant by this project.}\label{tab:all-sources}
\begin{xtabular}{@{}p{3.35cm} r p{2.85cm}@{}}
\raggedright TheoremDB & 3,909 & \raggedright Selected packets CC0-1.0; references keep their own terms \tabularnewline
\raggedright AIM Problem Lists & 3,470 & \raggedright Editorially moderated; per-list terms \tabularnewline
\raggedright ICM Conjectures & 2,898 & \raggedright Attribution retained; consult original proceedings \tabularnewline
\raggedright Erd\H{o}s Problems (Thomas Bloom) & 634 & \raggedright See site; statements paraphrased with attribution \tabularnewline
\raggedright Formal Conjectures & 459 & \raggedright Code Apache-2.0; text CC BY 4.0; third-party terms kept \tabularnewline
\raggedright Optimization Constants in Mathematics --- Davis, Ivanisvili \& Tao & 115 & \raggedright Apache-2.0 repository; pages and bound tables attributed \tabularnewline
\raggedright Open Quantum Problems & 114 & \raggedright Attribution retained; consult source for reuse terms \tabularnewline
\raggedright AlphaEvolve Repository of Problems --- Georgiev et al. & 59 & \raggedright Apache-2.0 repository; pages attributed \tabularnewline
\raggedright Epoch AI --- FrontierMath: Open Problems & 49 & \raggedright CC BY 4.0 (Epoch AI); pages and prompts attributed \tabularnewline
\raggedright Tableaux --- A notebook of open problems & 48 & \raggedright CC BY 4.0 (D. R. Worley); HTML to Markdown, attributed \tabularnewline
\raggedright Queueing, Predictions, and LLMs --- Mitzenmacher \& Shahout & 42 & \raggedright CC BY 4.0; attributed adaptation, HTML to Markdown \tabularnewline
\raggedright The Open Problems Project (computational geometry) & 38 & \raggedright Academic use with attribution \tabularnewline
\raggedright Community consensus & 34 & \raggedright Original wording \tabularnewline
\raggedright COLT Open Problems --- supplied archive & 30 & \raggedright Authors keep copyright; papers and audit notes attributed \tabularnewline
\raggedright Tianyuan Workshop --- Open Problems in Mathematical Logic & 22 & \raggedright CC BY 4.0; attributed adaptation, HTML to Markdown \tabularnewline
\raggedright Jacobs --- Twenty Open Problems in Quantum Control & 20 & \raggedright Author attribution and references kept; per-paper terms \tabularnewline
\raggedright Anton Petrunin --- A Gallery of Geometry Problems & 15 & \raggedright CC0 1.0 (A. Petrunin); independent summaries of v2 \tabularnewline
\raggedright Randomstrasse101 --- Open Problems of 2025 & 15 & \raggedright CC BY 4.0; Bandeira et al.; HTML to Markdown, attributed \tabularnewline
\raggedright Aaronson --- Quantum Query Complexity & 12 & \raggedright Author attribution and references kept; per-paper terms \tabularnewline
\raggedright Oberwolfach 2025 --- Analytic Number Theory Problem Session & 11 & \raggedright CC BY-SA 4.0; compiled by T. F. Bloom; share-alike kept \tabularnewline
\raggedright FOCS 2024 --- Quantum Learning & 10 & \raggedright Questions summarized with attribution; slides linked \tabularnewline
\raggedright Simons Institute --- Lattices '22 problems & 10 & \raggedright Summarized with attribution to the Lattices Program \tabularnewline
\raggedright Dagstuhl 25041 --- Graph Classes and Algorithmic Complexity & 8 & \raggedright CC BY 4.0; attributed editorial adaptation \tabularnewline
\raggedright Jo\~ao Ara\'ujo --- A survey on varieties generated by small semigroups and a companion website & 8 & \raggedright Brief editorial summaries; article keeps its own license \tabularnewline
\raggedright Wikipedia --- Computer science & 7 & \raggedright CC BY-SA 4.0; adapted selections, contributors attributed \tabularnewline
\raggedright Clay Mathematics Institute --- Millennium Prize Problems & 6 & \raggedright Statements paraphrased; official formulations linked \tabularnewline
\raggedright Wikipedia --- Loop theory and quasigroup theory & 6 & \raggedright CC BY-SA 4.0; adapted selections, contributors attributed \tabularnewline
\raggedright Bertrand Clarke --- A Cheat Sheet for Bayesian Prediction & 5 & \raggedright Brief editorial summaries; article keeps its own license \tabularnewline
\raggedright Wikipedia --- Astronomy & 5 & \raggedright CC BY-SA 4.0; adapted selections, contributors attributed \tabularnewline
\raggedright Wikipedia --- Biology & 5 & \raggedright CC BY-SA 4.0; adapted selections, contributors attributed \tabularnewline
\raggedright Wikipedia --- Mathematics & 5 & \raggedright CC BY-SA 4.0; adapted selections, contributors attributed \tabularnewline
\raggedright Wikipedia --- Physics & 5 & \raggedright CC BY-SA 4.0; adapted selections, contributors attributed \tabularnewline
\raggedright Dagstuhl 25081 --- Semirings in Database Theory & 4 & \raggedright CC BY 4.0; attributed editorial adaptation \tabularnewline
\raggedright JMLR --- Pre-trained Gaussian Processes for Bayesian Optimization & 4 & \raggedright CC BY 4.0; attributed editorial adaptation \tabularnewline
\raggedright Open Problems in Mathematical Biology --- Vittadello \& Stumpf & 4 & \raggedright CC BY-SA 4.0; attributed adaptation, HTML to Markdown \tabularnewline
\raggedright O\u{g}uz \c{S}avk --- A survey of the homology cobordism group & 4 & \raggedright Brief editorial summaries; article keeps its own license \tabularnewline
\raggedright Wikipedia --- Geoscience & 4 & \raggedright CC BY-SA 4.0; adapted selections, contributors attributed \tabularnewline
\raggedright Wikipedia --- Information theory & 4 & \raggedright CC BY-SA 4.0; adapted selections, contributors attributed \tabularnewline
\raggedright Wikipedia --- Neuroscience & 4 & \raggedright CC BY-SA 4.0; adapted selections, contributors attributed \tabularnewline
\raggedright Alessandro Carlotto --- The general relativistic constraint equations & 3 & \raggedright Brief editorial summaries; article keeps its own license \tabularnewline
\raggedright Dagstuhl --- Quantum Error Correction Meets ZX-Calculus & 3 & \raggedright CC BY 4.0; section authors attributed \tabularnewline
\raggedright FDIG 2025 --- Open Problems in Information Geometry & 3 & \raggedright Independent short summaries and pointers; no blanket license \tabularnewline
\raggedright Liouville Quantum Gravity --- Holden \& Sun & 3 & \raggedright CC BY 4.0; attributed adaptation, HTML to Markdown \tabularnewline
\raggedright Robert Benedetto --- Current trends and open problems in arithmetic dynamics & 3 & \raggedright Brief editorial summaries; article keeps its own license \tabularnewline
\raggedright Universality in Computable Dynamical Systems --- Old and New & 3 & \raggedright Independent short summaries and pointers; no blanket license \tabularnewline
\raggedright Wikipedia --- Chemistry & 3 & \raggedright CC BY-SA 4.0; adapted selections, contributors attributed \tabularnewline
\raggedright Andrii Mironchenko --- Input-to-state stability of infinite-dimensional systems: recent results and open questions & 2 & \raggedright Brief editorial summaries; article keeps its own license \tabularnewline
\raggedright Armin Schikorra --- Invitation to H-systems in higher dimensions: known results, new facts, and related open problems & 2 & \raggedright Brief editorial summaries; article keeps its own license \tabularnewline
\raggedright Axial Algebras --- Questions and Conjectures & 2 & \raggedright Independent short summaries and pointers; no blanket license \tabularnewline
\raggedright Ewin Tang --- Quantum Learning Questions & 2 & \raggedright Author attribution retained; consult source for terms \tabularnewline
\raggedright Open Problem Garden & 2 & \raggedright Community wiki \tabularnewline
\raggedright SIAM Optimization --- Open Problems in Optimization & 2 & \raggedright Independent short summaries and pointers; no blanket license \tabularnewline
\raggedright Simons Workshop --- Linear Systems and Eigenvalue Problems & 2 & \raggedright Independent short summaries and pointers; no blanket license \tabularnewline
\raggedright Smale's problems for the 21st century & 2 & \raggedright Paraphrased with citation \tabularnewline
\raggedright Wikipedia --- Economics & 2 & \raggedright CC BY-SA 4.0; adapted selections, contributors attributed \tabularnewline
\raggedright Wikipedia --- Fair division & 2 & \raggedright CC BY-SA 4.0; adapted selections, contributors attributed \tabularnewline
\raggedright Wikipedia --- Statistics & 2 & \raggedright CC BY-SA 4.0; adapted selections, contributors attributed \tabularnewline
\raggedright arXiv preprints & 2 & \raggedright Per-paper license; metadata via arXiv API \tabularnewline
\raggedright JMLR --- Fast Rates in PAC-Bayes Meta-Learning & 1 & \raggedright CC BY 4.0; attributed editorial adaptation \tabularnewline
\raggedright Kirby's problem list in low-dimensional topology & 1 & \raggedright Paraphrased with citation \tabularnewline
\raggedright Nathan Segerlind --- The complexity of propositional proofs & 1 & \raggedright Brief editorial summaries; article keeps its own license \tabularnewline
\raggedright Xavier Blanc --- The crystallization conjecture: a review & 1 & \raggedright Brief editorial summaries; article keeps its own license \tabularnewline
\end{xtabular}
}

\section{Curation rubric}\label{app:rubric}

This appendix records the rules used to admit, tier, score and select benchmark items. Scores are ordinal curator judgments, not expert consensus and not predictions of model success.

\paragraph{Hard filters.}
A candidate enters \redisc{} only if (i)~it was demonstrably a public research question before its resolution, with a traceable source; (ii)~the located solution settles the same quantifiers and scope as the posed target, so that a special case, a ``sufficiently large $n$'' result or a strengthened hypothesis does not count; (iii)~no earlier solution was located before its round's window (after 2026-06-30 for R items, from 2026-01-01 for E and N items); (iv)~the solution is not a restatement of an old theorem, a withdrawn or retracted paper, or an announcement without proof; and (v)~equivalent forms, parameter variants and direct corollaries of one construction are merged into one problem family. An open item must be open in both directions at the status check, precisely posed and not a duplicate or special case of another item. Every Lean statement records its scope differences from the original question (for instance, a universe restriction that does not affect the answer).

\paragraph{Split groups.}
Items that share a solution paper, a public release or a source are bound to one split group and should be reported as a group: the OpenAI ``ten advances'' release (N06, N16, N19, N20, N31, N35, N46), the Epoch AI FrontierMath Erd\H{o}s release (N10, N11, N12, N41), Benkoski--Erd\H{o}s 1974 (N07, O66), arXiv:2609.20094 (R28, R29) and one paper solving Erd\H{o}s \#987 and \#990 (E14, E15). All other items form singleton groups.

\paragraph{Evidence tiers.}
Tier~A (85; R 14, E 23, N 48): a finite certificate was independently recomputed with exact arithmetic, or a complete Lean proof of the answered statement was compiled locally, in most cases an AI-assisted formalization; 75 Tier~A items rest on a Lean proof and 10 on a certificate. Four Tier~A items (R26, N13, N49, N50) are conditional, each with the condition stated in its tier reason, and are excluded from \hq{}. Tier~B (21): a complete solution is corroborated by the problem list's acceptance, publication or an independent solution, but the full proof was not independently verified. Tier~C (4: R22, R24, R25, R30): a solution claim whose verification is pending.

\paragraph{Scores.}
Each item carries ordinal 1--5 curator scores for importance (1: no discernible research motivation; 3: a traceable, specialised research question; 5: a recognized central conjecture), statement clarity, formalization cost and proof cost. Importance is 3/4/5 for 22/49/39 solved items (mean 4.15) and 6/45/39 open items (mean 4.37). Scores are not combined into a composite except by the \hq{} ranking below.

\paragraph{Dates.}
Each solved item stores the earliest located public solution date and its basis (arXiv v1, dated public proof post, problem-list edition, repository commit or manuscript date). The audit read version histories, earlier PDFs, forum threads and pinned commits, but it was not an exhaustive priority search; a later publication, simplification or formalization does not reset a date. Located dates run starting from 2026-01-05.

\paragraph{Leakage controls in the statements.}
Lean files of solved items contain no answers, upstream declaration names or live links that display status. Citations of erdosproblems.com, OEIS and Green's problem list are given as literature references rather than live pages, Wikipedia links point to a pre-solution revision, and 16 Kourovka items cite a pre-solution arXiv version of the notebook. A second pass over the 110 solved files removed semantic hints (declarative titles became questions; background sentences pointing to the solution route were deleted). In the solver packet, items are renamed to P001--P200 in a salted order, file headers are unified and status-revealing source lines are removed.

\paragraph{\hq{} selection.}
For \hq{}, we have several more filters compared to \bench{}. Hard filters for all items: the statement compiles and passes the checker and probes, formalization confidence is high, no must-fix issue, and if both review checks raised issues are all fixed; a newly written statement must have had its primary-source wording verified. Solved items must be unconditional Tier~A with importance $\geq 4$ (importance 3 only if human-solved, post-cutoff and reviewed both passed). Open items need importance $\geq 4$, clarity 5, both passed reviews, an unrepaired statement, an answer not easily guessed (excluding O71) and open status at the check. The most frequent failures were open-item reviews other than both passes (22), Tier~B (21), open clarity below 5 (19), unfixed issues from both checks (16) and importance below 4 (21 in total); 55 solved and 48 open items qualified. The ranking adds points for review outcome (both passes $+2$, one passes $+1$), importance ($5$: $+2$; $3$: $-1$), a full Lean reference ($+2$; certificate $+1$), recency (after 2026-07-01 $+2$; April--June $+1$), proof cost $\geq 4$ ($+1$) and, for open items, a verbatim upstream statement ($+1.5$), a finite-certificate direction ($+1$) and priority; it subtracts for unread primary sources, unfixed issues, repair rounds and many scope differences. Greedy selection enforces one item per solution paper or problem family, two per public release (three for the OpenAI release of seven unrelated problems), at most 16 Erd\H{o}s and 6 Kourovka solved items, at most 4 Erd\H{o}s open items, at most 8 solved and 2 open items per area, and at most 21 solved items whose key mathematics came from OpenAI models, so that 19 remain for evaluating OpenAI models. Two items were excluded by hand (N03, N24; Appendix~\ref{app:exclusions}).

\section{Lean statements and checks}\label{app:lean}

\paragraph{Corpus and platform.}
Corpus Lean attachments (1,961 for 1,946 problems) are display-only: no Lean is executed at export time, the recorded lexical \texttt{sorry}/\texttt{admit}/\texttt{axiom} scans are not a proof checker, and source-side ``verified'' or ``solved'' labels are kept as dated claims. Community submissions on the platform are checked separately (profile \texttt{lean4-core-4.19.0-v1}: Lean 4.19.0 with \texttt{Std} only, the theorem built from the stored target, only \texttt{propext}, \texttt{Classical.choice} and \texttt{Quot.sound} allowed, a fresh networkless container per check); a successful check attaches a receipt to the submission and never changes a problem's status. 

\paragraph{Benchmark package.}
The 200 statements form one Lake project pinned to Lean~4.33.1, Mathlib \texttt{0df444a} and Formal Conjectures \texttt{89294ea}, of which only the two utility libraries are vendored; a clean \texttt{lake build} succeeds. Each file declares exactly one theorem \texttt{FrontierBench.<ID>.main}, and \texttt{sorry} may appear only in the answer placeholder and in the proof of \texttt{main}. Besides the checks of Section~\ref{sec:bench-design}, automated probes run on \texttt{main} with the correct and a wrong answer for solved yes/no items, with both answers for open yes/no items and with the correct answer for value items; no probe proves or refutes a statement. As corroboration of fidelity, 50 solved items had their answered \texttt{main} proved in this environment from public Lean proofs or recomputed certificates.

\paragraph{Reviews.}
Two AI reviewers per item worked independently: one compared the statement clause by clause with the original source, the other tried to prove or refute it cheaply through formalization loopholes. After at most two repair rounds (178 items needed none, 20 one, 2 two), the verdicts were 139 both passes, 38 with one minor problem, and 23 with two minor problems, with no must-fix issue; formalization confidence is high for 195 items and medium for 5.

\section{Evaluation details}\label{app:eval}

\begin{table*}[t]
\centering
\scriptsize
\setlength{\tabcolsep}{3pt}
\begin{tabular}{@{}l l c r r c p{6.6cm}@{}}
\toprule
Model & Prot. & Cond. & Median min & .lean & Open acc. & End reasons \\
\midrule
Claude Sonnet 5.5 & nat. & A & 0.3 & 3 & 0 & \raggedright agent stopped 49, agent error 1 \tabularnewline
 &  & B & 0.4 & 18 & 0 & \raggedright agent stopped 48, wall clock 2 \tabularnewline
 &  & C & 0.4 & 19 & 0 & \raggedright agent stopped 48, wall clock 2 \tabularnewline
Claude Sonnet 5.5 & cont. & A & 5.1 & 33 & 0 & \raggedright stalled 41, wall clock 4, submitted 4, agent stopped 1 \tabularnewline
 &  & B & 17.7 & 47 & 0 & \raggedright stalled 45, submitted 3, agent stopped 2 \tabularnewline
 &  & C & 35.9 & 46 & 0 & \raggedright stalled 27, submitted 23 \tabularnewline
\addlinespace
GPT-6.1-sol & nat. & A & 3.1 & 25 & 0 & \raggedright agent stopped 50 \tabularnewline
 &  & B & 3.6 & 11 & 0 & \raggedright agent stopped 50 \tabularnewline
 &  & C & 4.0 & 28 & 0 & \raggedright agent stopped 50 \tabularnewline
GPT-6.1-sol & cont. & A & 180.0 & 50 & 0 & \raggedright wall clock 40, submitted 8, agent stopped 2 \tabularnewline
 &  & B & 180.0 & 47 & 0 & \raggedright wall clock 42, submitted 6, agent stopped 2 \tabularnewline
 &  & C & 101.1 & 50 & 1 & \raggedright submitted 29, wall clock 18, agent stopped 3 \tabularnewline
\addlinespace
GPT-6-luna & nat. & A & 1.7 & 0 & 0 & \raggedright agent stopped 50 \tabularnewline
 &  & B & 3.1 & 0 & 0 & \raggedright agent stopped 50 \tabularnewline
 &  & C & 4.1 & 6 & 0 & \raggedright agent stopped 49, wall clock 1 \tabularnewline
GPT-6-luna & cont. & A & 180.0 & 36 & 0 & \raggedright wall clock 41, agent stopped 6, submitted 2, disk quota 1 \tabularnewline
 &  & B & 180.0 & 33 & 0 & \raggedright wall clock 45, agent stopped 4, submitted 1 \tabularnewline
 &  & C & 180.0 & 41 & 0 & \raggedright wall clock 32, submitted 13, agent stopped 4, disk quota 1 \tabularnewline
\addlinespace
Kimi-K3 & nat. & A & 30.3 & 16 & 0 & \raggedright agent stopped 50 \tabularnewline
 &  & B & 26.7 & 10 & 0 & \raggedright agent stopped 50 \tabularnewline
 &  & C & 26.6 & 28 & 0 & \raggedright agent stopped 48, wall clock 2 \tabularnewline
Kimi-K3 & cont. & A & 179.2 & 24 & 0 & \raggedright agent stopped 38, wall clock 8, submitted 3, stalled 1 \tabularnewline
 &  & B & 179.2 & 22 & 0 & \raggedright agent stopped 31, wall clock 12, stalled 5, submitted 2 \tabularnewline
 &  & C & 179.1 & 32 & 0 & \raggedright agent stopped 25, submitted 13, wall clock 11, stalled 1 \tabularnewline
\addlinespace
GLM-5.3 & nat. & A & 90.0 & 17 & 0 & \raggedright agent stopped 44, wall clock 4, agent error 2 \tabularnewline
 &  & B & 104.7 & 11 & 0 & \raggedright agent stopped 49, wall clock 1 \tabularnewline
 &  & C & 102.2 & 28 & 0 & \raggedright agent stopped 47, wall clock 2, agent error 1 \tabularnewline
GLM-5.3 & cont. & A & 180.0 & 22 & 0 & \raggedright wall clock 26, agent stopped 24 \tabularnewline
 &  & B & 179.4 & 25 & 0 & \raggedright agent stopped 30, wall clock 18, submitted 2 \tabularnewline
 &  & C & 179.6 & 38 & 1 & \raggedright wall clock 22, submitted 14, agent stopped 14 \tabularnewline
\addlinespace
DeepSeek-V4.1-Flash & nat. & A & 22.2 & 8 & 0 & \raggedright agent stopped 48, agent error 2 \tabularnewline
 &  & B & 22.0 & 16 & 0 & \raggedright agent stopped 50 \tabularnewline
 &  & C & 14.2 & 22 & 0 & \raggedright agent stopped 50 \tabularnewline
DeepSeek-V4.1-Flash & cont. & A & 179.2 & 20 & 0 & \raggedright agent stopped 23, wall clock 14, stalled 10, submitted 3 \tabularnewline
 &  & B & 179.2 & 22 & 0 & \raggedright agent stopped 31, wall clock 10, stalled 6, submitted 3 \tabularnewline
 &  & C & 179.1 & 29 & 1 & \raggedright agent stopped 21, submitted 15, wall clock 7, stalled 7 \tabularnewline
\addlinespace
Qwen3.8-27B & nat. & A & 180.1 & 21 & 0 & \raggedright wall clock 46, agent error 3, agent stopped 1 \tabularnewline
 &  & B & 180.0 & 11 & 0 & \raggedright wall clock 30, agent error 13, agent stopped 7 \tabularnewline
 &  & C & 180.0 & 18 & 0 & \raggedright wall clock 38, agent error 6, agent stopped 5, api error 1 \tabularnewline
Qwen3.8-27B & cont. & A & 180.0 & 15 & 0 & \raggedright wall clock 50 \tabularnewline
 &  & B & 180.0 & 23 & 0 & \raggedright wall clock 50 \tabularnewline
 &  & C & 180.0 & 30 & 0 & \raggedright wall clock 50 \tabularnewline
\addlinespace
\bottomrule
\end{tabular}

\caption{Per-run details on \hq{} (50 items per condition, one attempt each; 180-minute budget). Median min = median attempt wall-clock minutes; .lean = attempts that left a formal submission; Open acc.\ = open items accepted by both checkers, before review.}
\label{tab:eval-details}
\end{table*}

\paragraph{Isolation.}
Every command of an attempt runs in a freshly built root file system that holds only the attempt's work directory, read-only prebuilt dependencies and the Lean toolchain, under its own unprivileged user ID with all capabilities dropped, private namespaces (network included) and a seccomp filter that forbids sockets, new namespaces, mounts and \texttt{ptrace}. Per attempt, cgroups enforce 5 CPU cores, 24\,GB of memory and 4,096 processes, and the work directory and \texttt{/tmp} may hold at most 10\,GB. The agent CLI runs in an identical jail with a loopback-only network whose only exits are a model API gateway (which routes every request, auxiliary calls included, to the evaluated model, keeps server-side web tools only in C and removes code execution and remote MCP tools), an MCP server offering \texttt{check\_submission} and the retrieval tools of B or C, and, in C only, a web proxy that refuses private addresses. A 20-item jail check passes on every node.

\paragraph{Harness and protocols.}
The CLIs run with their own system prompt, tools, context management and default reasoning settings. Claude Code's output limit was raised to 128k tokens, since GLM-5.3 and Qwen3.8-27B often think longer than its default of 32k for unknown models; Qwen3.8-27B was served with sglang (262k-token context). The task prompt contains the packet README, the problem file, the submission rules and the budget. \texttt{check\_submission} applies the grader's static rules, compares the definitions and the statement of \texttt{main} with the original, compiles and lists the axioms of \texttt{main}, without consulting the reference answer. Under continuation, a session that ends with at least one minute left and no passing submission is resumed (\texttt{claude --resume}, \texttt{codex exec resume}) with a message stating the remaining time, the submission status and concrete tool work to try; three consecutive rounds without a successful model call end the attempt as an agent error, and the models are not told about the stall rule.

\paragraph{Frozen retrieval (B).}
The cutoff, 2026-01-06, is the same for all items and announced to the model. arXiv returns papers first submitted before the cutoff, in their latest pre-cutoff version; Wikipedia, MathOverflow and Math.SE return the last pre-cutoff revision (for Stack Exchange, only questions, answers and comments posted before the cutoff); OpenAlex shows only journal or proceedings records, because merged preprint records can carry post-cutoff abstracts; other sites are refused. Since search engines index today's text, every hit is re-verified against the displayed historical text.

\paragraph{Grading.}
Grading runs on the evaluator's machine with the evaluator-only answers. The compile steps run in a networkless jail whose mounts are read-only except the step's build directory, so that code executed at compile time cannot alter the reference, the grader, the answers or Mathlib; the comparison reads the compiled modules rather than parsing source. Comparator, which exports both projects with \texttt{lean4export} and replays the full export, is pinned and ported to Lean~4.33, with a fix that also counts structures referenced by projections, without which its own projection attack would pass. In C, every fetched URL is normalized (arXiv ID, DOI, repository, problem page) and compared with the item's located solution sources, to flag attempts that fetched one.

\paragraph{Models and cutoffs.}
We used Claude Code and Codex as the agent harness. Claude Sonnet 5.5, Kimi-K3, GLM-5.3 and DeepSeek-V4.1-Flash were reached through an Anthropic-compatible bridge, Qwen3.8-27B through sglang's Messages endpoint, and GPT-6.1-sol and GPT-6-luna through a Responses-API bridge. Grading used the lineage of each provider for the two OpenAI models and Claude Sonnet 5.5, and no lineage exclusion for the others, whose own solution credits do not occur in \hq{}. Cutoffs were 2026-06-30 for the Anthropic and OpenAI models, and the public release dates for the open-weight models (Kimi-K3 2026-06-13, Qwen3.8-27B 2026-08-05, GLM-5.3 2026-08-25, DeepSeek-V4.1-Flash 2026-09-10); the late release of DeepSeek-V4.1-Flash leaves only 5 clean items. Table~\ref{tab:eval-details} gives per-run details; all attempts used evalkit-0.3.

\section{Dataset format and licensing}\label{app:schema}

\paragraph{Format.}
Problems in the corpus are unified into a universal JSON Schema that documents every field, a validation report, checksums, an input manifest and other necessary information. Unknown values are null, and missing data is never converted into invented statements or proofs.
Figure~\ref{fig:record-example} shows an abridged problem record. It keeps the source's own license wording, an MSC code with its mapping basis, a dated status, and an upstream Lean statement.

\begin{figure}[t]
\scriptsize
\begin{verbatim}
{"id": "problem:erdos-7",
 "record_kind": "problem",
 "title": "Erdős #7 · number theory /
     covering systems",
 "problem_statement": "Is there a distinct covering
     system all of whose moduli are odd?",
 "background": "Asked by Erdős and Selfridge ...",
 "field": {"main": "math",
     "slug": "number-theory", ...},
 "msc2020": [{"code": "11",
     "basis": "community-map",
     "evidence": "Inferred from community field:
     number-theory"}],
 "status": "open",
 "source": {"key": "erdos-problems",
     "type": "problem-list",
     "url": "https://www.erdosproblems.com/7",
     "license": "See site; statements paraphrased
     with attribution", ...},
 "provenance": {"imported_at": "2026-09-08",
     "status_as_of": "2026-09-08",
     "has_full_text": true,
     "source_record":
       "https://www.erdosproblems.com/7",
     "sha256": "71fbda4a...698e7172", ...},
 "references": [{"kind": "website",
     "citation":
       "Thomas Bloom (ed.), Erdős Problems #7",
     "url": "https://www.erdosproblems.com/7", ...},
     {"kind": "formalization", ...}, ...],
 "formalizations": [{"language": "Lean 4",
     "format": "source-file",
     "commit": "b82b08fa...",
     "toolchain": "leanprover/lean4:v4.33.1",
     "license": "Apache-2.0; original file ...",
     "has_sorry_or_admit": true,
     "verification_status": "not-checked",
     "code": "... @[category research open, AMS 11]
       theorem erdos_7 : answer(sorry) ↔
         ∃ (C : StrictCoveringSystem ℤ), ∀ i,
           ¬ C.moduli i ≤ Ideal.span {2} ∧
           C.moduli i ≠ ⊤ := by
         sorry ...", ...}],
 "proof": null, "proof_status": "not-recorded",
 "lean4_proof": null,
 "lean4_proof_status": "not-recorded",
 "quality_flags": [
     "lean-statement-from-linked-source-file",
     "formalization-contains-placeholder",
     "formalization-not-compiled-during-export"],
 "schema_version": "1.0.0",
 "snapshot_at": "2026-09-10T06:58:35.349Z", ...}
\end{verbatim}
\caption{An abridged problem record (Erd\H{o}s Problem~\#7) from the \openfrontiers{} collection.}
\label{fig:record-example}
\end{figure}

\paragraph{Licensing and intended use.}
\openfrontiers{} is a mixed-source dataset and grants no new uniform license over third-party works. Each item keeps its source's reuse wording in \texttt{source.license} and each Lean attachment its own terms in \texttt{formalizations[].license}; missing or ambiguous terms mean that no license was established. Terms range from CC0 and Apache-2.0 to share-alike (CC BY-SA: AIM Problem Lists, Wikipedia) and non-commercial ones (the bundled MSC 2020 taxonomy, CC BY-NC-SA 4.0), so users who need a permissive subset must filter by source key. The corpus is intended for locating research problems and evaluating research-level reasoning, not as permissively licensed training data.

\section{Exclusions and changes}\label{app:exclusions}

We conduct multiple rounds of benchmark filtering during construction.
For example, re-reviewing the earlier version removed R09 (Kourovka 11.9; the answer was already in the literature) and R11 (Kourovka 13.23; the audit found a non-inner automorphism of the group constructed in the solution preprint), and moved four open items resolved in 2026 to \redisc{}: N01 (Beck--Fiala), N02 (Kourovka 20.76), N03 (Tu--Deng) and N04 (irrationality of $\zeta(5)$); IDs are not reused. The overturn review rejected five new candidates, replaced from the reserve: Erd\H{o}s \#986 and the rank-two Poisson conjecture for overlap with admitted items, Vaught's conjecture because its upstream cardinal form follows trivially from the continuum hypothesis, and two special cases or duplicates of admitted items ($2^x, 3^x \in \mathbb{Z} \Rightarrow x \in \mathbb{Z}$; Green's problem 28). Two items were excluded from \hq{} by hand: N03 (two repair rounds, and the original Tu--Deng wording is known only through later papers) and N24 (its reduction rests on a single unconfirmed preprint, and the answered statement has no Lean proof). The open status will be regularly updated over time as more previously open problems are gradually resolved.

\section{Use of AI assistants}\label{app:ai-use}
We mainly used generative AI agents to locate candidate problems and solution materials and audit their status, write and review the Lean statements of \bench{} (Section~\ref{sec:bench-design}), help build the evaluation harness, grader, and statistics scripts, and polish this paper.

\end{document}